\documentclass[letterpaper]{article} 
\usepackage{aaai24}  
\usepackage{times}  
\usepackage{helvet}  
\usepackage{courier}  
\usepackage[hyphens]{url}  
\usepackage{graphicx} 
\usepackage{natbib}  
\usepackage{caption} 
\usepackage{algorithm}
\usepackage{algorithmic}

\usepackage{subcaption}
\usepackage{csquotes}
\usepackage{booktabs}
\usepackage{multirow}
\usepackage{amsmath}
\usepackage{tikz}
\usepackage{todonotes}
\usepackage[dont-mess-around]{fnpct}

\usepackage{todonotes}

\title{Low-Resource Authorship Style Transfer: Can Non-Famous Authors Be Imitated?}
\author {
    Ajay Patel\textsuperscript{\rm 1},
    Nicholas Andrews\textsuperscript{\rm 2},
    Chris Callison-Burch\textsuperscript{\rm 1}
}
\affiliations {
    \textsuperscript{\rm 1}University of Pennsylvania\\
    \textsuperscript{\rm 2}Johns Hopkins University\\
    ajayp@upenn.edu, noa@jhu.edu, ccb@upenn.edu
}

\begin{document}

\maketitle

\newcommand{\uarembeddingresultstable}{
    \begin{table*}
    \centering
    \setlength{\tabcolsep}{1.8pt}
    \fontsize{9}{11}\selectfont
    \begin{tabular}{lrrrr|rrrr|rrrr}
    \toprule[\heavyrulewidth]
    Method ~~~ & \multicolumn{4}{c|}{Random}& \multicolumn{4}{c|}{Single}& \multicolumn{4}{c}{Diverse}\\
           & \multicolumn{1}{c}{\textsc{Away}} & \multicolumn{1}{c}{\textsc{Towards}} & \multicolumn{1}{c}{\textsc{Sim}} & \multicolumn{1}{c|}{\textsc{Joint}} & \multicolumn{1}{c}{\textsc{Away}} & \multicolumn{1}{c}{\textsc{Towards}} & \multicolumn{1}{c}{\textsc{Sim}} & \multicolumn{1}{c|}{\textsc{Joint}} & \multicolumn{1}{c}{\textsc{Away}} & \multicolumn{1}{c}{\textsc{Towards}} & \multicolumn{1}{c}{\textsc{Sim}} & \multicolumn{1}{c}{\textsc{Joint}} \\ \hline
    $\textsc{Copy}_{\textsc{src}}$   & 0.00 & 0.00 & 1.00 & 0.00 & 0.00 & 0.00 & 1.00 & 0.00 & 0.00 & 0.00 & 1.00 & 0.00 \\
    $\textsc{Copy}_{\textsc{tgt}}$    & 1.00 & 1.00 & 0.00 & 0.00 & 1.00 & 1.00 & 0.00 & 0.00 & 1.00 & 1.00 & 0.00 & 0.00 \\ \hline
    \textsc{Capi}   & 0.42 & 0.02 & 0.89 & 0.17 & 0.56 & 0.01 & 0.93 & 0.07 & 0.41 & 0.01 & 0.87 & 0.08 \\
    \textsc{Cont}   & 0.20 & 0.01 & 0.91 & 0.15 & 0.22 & 0.02 & 0.97 & 0.16 & 0.21 & 0.01 & 0.93 & 0.13 \\
    \textsc{Synm}   & 0.23 & 0.02 & 0.92 & 0.17 & 0.22 & 0.01 & 0.95 & 0.10 & 0.17 & 0.01 & 0.91 & 0.07 \\
    \textsc{Punc}   & 0.23 & 0.02 & 0.93 & 0.24 & 0.25 & 0.02 & 0.97 & 0.19 & 0.26 & 0.02 & 0.90 & 0.18 \\
    \textsc{Emoj}   & 0.27 & 0.04 & 0.93 & 0.25 & 0.29 & 0.06 & 0.95 & 0.27 & 0.27 & 0.02 & 0.93 & 0.17 \\ \hline
    $\textsc{Para}_{\textsc{Neu}}$   & 0.78 & 0.01 & 0.58 & 0.05 & 0.91 & 0.01 & 0.60 & 0.05 & 0.87 & 0.05 & 0.53 & 0.18 \\
    $\textsc{Para}_{\textsc{Div}}$   & 0.75 & 0.01 & 0.69 & 0.10 & 0.91 & 0.02 & 0.71 & 0.11 & 0.83 & 0.04 & 0.70 & 0.18 \\
    \textsc{Ling}   & 0.60 & 0.06 & 0.85 & 0.32 & 0.71 & 0.06 & 0.88 & 0.25 & 0.57 & 0.03 & 0.81 & 0.19 \\
    \textsc{Bert}   & 0.22 & 0.02 & 0.72 & 0.13 & 0.29 & 0.01 & 0.69 & 0.10 & 0.30 & 0.01 & 0.60 & 0.08 \\
    $\textsc{Strap}_{p=0.0}$   & 0.98 & 0.02 & 0.16 & 0.05 & 1.00 & 0.00 & 0.23 & 0.00 & 0.97 & 0.02 & 0.22 & 0.04 \\
    $\textsc{Strap}_{p=0.6}$   & 0.99 & 0.02 & 0.08 & 0.04 & 1.00 & 0.00 & 0.13 & 0.01 & 0.97 & 0.02 & 0.11 & 0.03 \\
    $\textsc{Strap}_{p=0.9}$   & 0.99 & 0.02 & 0.05 & 0.03 & 1.00 & 0.00 & 0.08 & 0.00 & 0.97 & 0.01 & 0.05 & 0.01 \\ \hline
    $\textsc{Styll}_{\text{GPT-3}}$  & 0.78 & 0.07 & 0.45 & 0.23 & 0.91 & 0.11 & 0.48 & 0.29 & 0.87 & 0.12 & 0.44 & 0.30 \\
    $\textsc{Styll}_{\text{BLOOM}}$  & 0.70 & 0.11 & 0.54 & \textbf{0.34} & 0.86 & 0.16 & 0.57 & \textbf{0.40} & 0.76 & 0.12 & 0.58 & \textbf{0.36} \\
    \bottomrule[\heavyrulewidth]
    \end{tabular}
    \caption{Automatic evaluation metrics for our set of baselines over three dataset variants in the UAR Embedding space. Our method \textsc{Styll} outperforms on the proposed \textsc{Joint} metric.}
    \label{table:uar-embedding-results}
    \end{table*}
}

\newcommand{\styleembeddingresultstable}{
    \begin{table*}[h]
    \centering
    \setlength{\tabcolsep}{1.8pt}
    \fontsize{9}{11}\selectfont
    \begin{tabular}{lrrrr|rrrr|rrrr}
    \toprule[\heavyrulewidth]
    Method ~~~ & \multicolumn{4}{c|}{Random}& \multicolumn{4}{c|}{Single}& \multicolumn{4}{c}{Diverse}\\
           & \multicolumn{1}{c}{\textsc{Away}} & \multicolumn{1}{c}{\textsc{Towards}} & \multicolumn{1}{c}{\textsc{Sim}} & \multicolumn{1}{c|}{\textsc{Joint}} & \multicolumn{1}{c}{\textsc{Away}} & \multicolumn{1}{c}{\textsc{Towards}} & \multicolumn{1}{c}{\textsc{Sim}} & \multicolumn{1}{c|}{\textsc{Joint}} & \multicolumn{1}{c}{\textsc{Away}} & \multicolumn{1}{c}{\textsc{Towards}} & \multicolumn{1}{c}{\textsc{Sim}} & \multicolumn{1}{c}{\textsc{Joint}} \\ \hline
    $\textsc{Copy}_{\textsc{src}}$   & 0.00 & 0.00 & 1.00 & 0.00 & 0.00 & 0.00 & 1.00 & 0.00 & 0.00 & 0.00 & 1.00 & 0.00 \\
    $\textsc{Copy}_{\textsc{tgt}}$    & 1.00 & 1.00 & 0.00 & 0.00 & 1.00 & 1.00 & 0.00 & 0.00 & 1.00 & 1.00 & 0.00 & 0.00 \\ \hline
    \textsc{Capi}   & 0.69 & 0.08 & 0.89 & 0.24 & 0.70 & 0.12 & 0.93 & 0.29 & 0.63 & 0.03 & 0.87 & 0.09 \\
    \textsc{Cont}   & 0.23 & 0.03 & 0.91 & 0.15 & 0.22 & 0.02 & 0.97 & 0.12 & 0.26 & 0.01 & 0.93 & 0.06 \\
    \textsc{Synm}   & 0.17 & 0.05 & 0.92 & 0.21 & 0.12 & 0.03 & 0.95 & 0.16 & 0.10 & 0.01 & 0.91 & 0.09 \\
    \textsc{Punc}   & 0.21 & 0.05 & 0.93 & 0.22 & 0.18 & 0.05 & 0.97 & 0.20 & 0.25 & 0.04 & 0.90 & 0.14 \\
    \textsc{Emoj}   & 0.17 & 0.03 & 0.93 & 0.14 & 0.12 & 0.03 & 0.95 & 0.14 & 0.12 & 0.01 & 0.93 & 0.09 \\ \hline
    $\textsc{Para}_{\textsc{Neu}}$   & 0.79 & 0.04 & 0.58 & 0.09 & 0.84 & 0.05 & 0.60 & 0.09 & 0.89 & 0.11 & 0.53 & 0.21 \\
    $\textsc{Para}_{\textsc{Div}}$   & 0.95 & 0.03 & 0.69 & 0.10 & 0.96 & 0.09 & 0.71 & 0.18 & 0.92 & 0.03 & 0.70 & 0.07 \\
    \textsc{Ling}   & 0.74 & 0.13 & 0.85 & \textbf{0.33} & 0.75 & 0.17 & 0.88 & 0.36 & 0.71 & 0.04 & 0.81 & 0.12 \\
    \textsc{Bert}   & 0.15 & 0.05 & 0.72 & 0.14 & 0.18 & 0.04 & 0.69 & 0.13 & 0.20 & 0.02 & 0.60 & 0.08 \\
    $\textsc{Strap}_{p=0.0}$   & 0.96 & 0.04 & 0.16 & 0.06 & 0.96 & 0.15 & 0.23 & 0.16 & 0.95 & 0.04 & 0.22 & 0.04 \\
    $\textsc{Strap}_{p=0.6}$   & 0.97 & 0.04 & 0.08 & 0.04 & 0.96 & 0.16 & 0.13 & 0.12 & 0.95 & 0.03 & 0.11 & 0.02 \\
    $\textsc{Strap}_{p=0.9}$   & 0.96 & 0.04 & 0.05 & 0.03 & 0.96 & 0.17 & 0.08 & 0.09 & 0.96 & 0.03 & 0.05 & 0.01 \\ \hline
    $\textsc{Styll}_{\text{GPT-3}}$  & 0.82 & 0.13 & 0.45 & 0.22 & 0.92 & 0.31 & 0.48 & \textbf{0.39} & 0.93 & 0.23 & 0.44 & \textbf{0.33} \\
    $\textsc{Styll}_{\text{BLOOM}}$  & 0.83 & 0.16 & 0.37 & 0.25 & 0.90 & 0.32 & 0.42 & 0.36 & 0.91 & 0.25 & 0.37 & 0.30 \\
    \bottomrule[\heavyrulewidth]
    \end{tabular}
    \caption{Automatic evaluation metrics for our set of baselines over three dataset variants in the Style Embedding space. Our method \textsc{Styll} outperforms baselines on the ``Single'' and ``Diverse'' variants on the proposed \textsc{Joint} metric. The \textsc{Ling} baseline outperforms on the ``Random'' variant where we expect authorship style transfer to be more undefined or outright impossible as discussed in-depth in Section 3. \textsc{Ling} has downsides as a handcrafted solution we discuss in Section 4, under $\textsc{Ling}$.}
    \end{table*}
}

\newcommand{\uarretrievalresultstable}{
    \begin{table*}
    \centering
    \fontsize{9}{11}\selectfont
    \begin{tabular}{llrrr|rrr|r}
    \toprule[\heavyrulewidth]
    & & \multicolumn{3}{c}{\underline{Source Author}} & \multicolumn{3}{c}{\underline{Target Author}} & \\\
    Dataset~~~ & Method ~~~ & \multicolumn{1}{c}{R@8} & \multicolumn{1}{c}{MRR} & \multicolumn{1}{c}{\textsc{Mean Rank}} & \multicolumn{1}{c}{R@8} & \multicolumn{1}{c}{MRR} & \multicolumn{1}{c}{\textsc{Mean Rank}} & \multicolumn{1}{c}{\textsc{Confusion}} \\ \hline
    \multirow{4}{*}{Random}
    & \textsc{Source}   & 0.73 & 0.55 & 11.67 & 0.00 & 0.00 & 49,553.51 & 0.00 \\
    & \textsc{Target}    & 0.00 & 0.00 & 54,454.28 & 0.60 & 0.43 & 425.93 & 1.00 \\ \cline{2-9} 
    & $\textsc{Styll}_{\text{GPT-3}}$  & 0.00 & 0.00 & 14,000.11 & 0.00 & 0.00 & 27,416.64 & \textbf{0.26} \\ 
    & $\textsc{Styll}_{\text{BLOOM}}$  & 0.07 & 0.03 & 8,913.84 & 0.00 & 0.00 & 25,262.64 & 0.19 \\ \hline
    \multirow{4}{*}{Single}
    & \textsc{Source}   & 0.53 & 0.50 & 46.07 & 0.00 & 0.00 & 10,329.16 & 0.01 \\
    & \textsc{Target}    & 0.00 & 0.00 & 12,211.86 & 0.67 & 0.46 & 518.40 & 0.93 \\ \cline{2-9} 
    & $\textsc{Styll}_{\text{GPT-3}}$  & 0.02 & 0.01 & 9,264.95 & 0.13 & 0.08 & 7,593.85 & \textbf{0.62} \\
    & $\textsc{Styll}_{\text{BLOOM}}$  & 0.00 & 0.00 & 5,816.22 & 0.12 & 0.06 & 4,144.24 & 0.60 \\ \hline
    \multirow{4}{*}{Diverse}
        & \textsc{Source}   & 0.40 & 0.32 & 2,569.87 & 0.00 & 0.00 & 37,122.09 & 0.05 \\
    & \textsc{Target}    & 0.00 & 0.00 & 53,932.73 & 0.67 & 0.49 & 257.93 & 1.00 \\ \cline{2-9} 
    & $\textsc{Styll}_{\text{GPT-3}}$  & 0.04 & 0.01 & 30,989.25 & 0.01 & 0.01 & 24,436.47 & \textbf{0.60} \\
    & $\textsc{Styll}_{\text{BLOOM}}$  & 0.06 & 0.06 & 22,419.18 & 0.00 & 0.00 & 19,515.24 & 0.53 \\
    \bottomrule[\heavyrulewidth]
    \end{tabular}
    \caption{Authorship identification performance with UAR embeddings over $|N| = 111,396$ authors with style transfer outputs from our method. \textsc{Styll} confuses the AID model over 50\% of the time on the ``Single'' and ``Diverse'' variants and forces the target author into the first 8 results 12-13\% of the time on the ``Single'' variant.}
    \label{table:uar-retrieval-results}
    \end{table*}
}

\newcommand{\postiveresultstable}{
    \begin{table*}[ht]
    \centering
    \fontsize{9}{11}\selectfont
    \begin{tabular}{p{5.2cm}p{1.7cm}p{4.2cm}p{4.5cm}}
    \toprule[\heavyrulewidth]
    Target Style Examples & Method & Input $~~\rightarrow~~$ Output & Analysis \\ \hline
        ``CRAP DID I MISS IT" \newline
        ``HES FALLING WITH STYLE!" \newline
        ``LOL DOESNT MATTER THEY SCORED" \newline
        \ldots 13 more examples &
        $\textsc{Styll}_{\text{BLOOM}}$ &
        ``That short throw was nice."$~~\rightarrow$~~\newline
        ``IT WAS A GOOD SHORT THROW" &
        Mimics the capitalization style of the target author and removes end mark punctuation.
    \\ \hline
        ``no way this dude can make this field goal" \newline
        ``lol he sure did" \newline
        ``more like bullet to the head" \newline
        \ldots 13 more examples &
        $\textsc{Styll}_{\text{GPT-3}}$ &
        ``WOOOOOOOOOOO. And the justin fuente era has officially started!"$~~\rightarrow$~~\newline
        ``justin fuente era officially started" &
        Transforms to lower case capitalization style and a terser style.
    \\ \hline
        ``oh crap..." \newline
        ``Sounds about right...  Georiga will likely move up to \#2" \newline
        ``Channeling madden..." \newline
        \ldots 13 more examples &
        $\textsc{Styll}_{\text{BLOOM}}$ &
        ``Lol I'm done already"$~~\rightarrow$~~\newline
        ``I'm done with this... I'm done" &
        Applies the use of ellipses, similar to the target author.
    \\ \hline
        ``Come to Dickson - I'm a man of my word" \newline
        ``You have my love - for the moment" \newline
        \ldots 14 more examples &
        $\textsc{Styll}_{\text{BLOOM}}$ &
        ``Holy shit! There's hope!"$~~\rightarrow$~~\newline
        ``Oh my God - there's hope" &
        Applies the use of an interruptive dash, similar to the target author.
    \\ \hline
        ``The [voyage of Jewish refugees]\newline(https://www.ushmm.org/wlc/en/\newline article.php?ModuleId=10005267) in the [St. Louis](http://www.bbc.com\newline /news/magazine-27373131) exemplifies this fact. \textbackslash n\textbackslash n Great fucking point, u/*******!" \newline
        \ldots 15 more examples &
        $\textsc{Styll}_{\text{GPT-3}}$ &
        ``After going through 4 submissions that broke the peak limiter this was a great one to land on. "$~~\rightarrow$~~\newline
        ``After reviewing [four submissions](https://www.reddit.com/r/\newline gifs/comments/3a0a6j/peak\_ \newline 
        limiter\_review\_4\_submissions\newline 
        \_that/)? this one was a standout" &
        The target author heavily provides links and references in Markdown. Our method follows this stylistic pattern, but the hyperlink it generates is of course a hallucination without the model having real-world knowledge of URLs.
    \\ \hline
        ``\#HEAD ON\#\textbackslash n\textbackslash n\#APPLY DIRECTLY TO THE FOREHEAD\#" \newline
        \ldots 15 more examples &
        $\textsc{Styll}_{\text{GPT-3}}$ &
        ``Happy cake day!\textbackslash n"$~~\rightarrow$~~\newline
        ``\#BIG HAPPY BIRTHDAY\#" &
        Mimics using pairs of "\#" around a capitalized message, a very unique indicator of the target author.
    \\ 
    \bottomrule[\heavyrulewidth]
    \end{tabular}
    \caption{Selected example outputs generated by \textsc{Styll} with analysis. More generations and examples of common failure modes can be found in Appendix H and I.}
    \label{table:positive-results}
    \end{table*}
}

\newcommand{\fullpostiveresultstable}{
    \begin{table*}[h]
    \centering
    \fontsize{9}{11}\selectfont
    \begin{tabular}{p{5.2cm}p{1.7cm}p{4.2cm}p{4.5cm}}
    \toprule[\heavyrulewidth]
    Target Style Examples & Method & Input $~~\rightarrow~~$ Output & Analysis \\ \hline
        ``CRAP DID I MISS IT" \newline
        ``HES FALLING WITH STYLE!" \newline
        ``LOL DOESNT MATTER THEY SCORED" \newline
        \ldots 13 more examples &
        $\textsc{Styll}_{\text{BLOOM}}$ &
        ``That short throw was nice."$~~\rightarrow$~~\newline
        ``IT WAS A GOOD SHORT THROW" &
        Mimics the capitalization style of the target author and removes end mark punctuation.
    \\ \hline
        ``no way this dude can make this field goal" \newline
        ``lol he sure did" \newline
        ``more like bullet to the head" \newline
        \ldots 13 more examples &
        $\textsc{Styll}_{\text{GPT-3}}$ &
        ``WOOOOOOOOOOO. And the justin fuente era has officially started!"$~~\rightarrow$~~\newline
        ``justin fuente era officially started" &
        Transforms to lower case capitalization style and a terser style.
    \\ \hline
        ``they keep hillbillies from getting dented? *sweet*" \newline
        ``of course, this is a *thing*" \newline
        ``*Never* forget your towel!" \newline
        \ldots 13 more examples &
        $\textsc{Styll}_{\text{GPT-3}}$ &
        ``SHUT UP PORTUGAL"$~~\rightarrow$~~\newline
        ``Please be *quiet*, Portugal" &
        Applies the use of asterisks (Markdown formatting) for emphasis instead of capitalization, similar to the target author.
    \\  \hline
        ``oh crap..." \newline
        ``Sounds about right...  Georiga will likely move up to \#2" \newline
        ``Channeling madden..." \newline
        \ldots 13 more examples &
        $\textsc{Styll}_{\text{BLOOM}}$ &
        ``Lol I'm done already"$~~\rightarrow$~~\newline
        ``I'm done with this... I'm done" &
        Applies the use of ellipses, similar to the target author.
    \\ \hline
        ``Come to Dickson - I'm a man of my word" \newline
        ``You have my love - for the moment" \newline
        \ldots 14 more examples &
        $\textsc{Styll}_{\text{BLOOM}}$ &
        ``Holy shit! There's hope!"$~~\rightarrow$~~\newline
        ``Oh my God - there's hope" &
        Applies the use of an interruptive dash, similar to the target author.
    \\ \hline
        ``The [voyage of Jewish refugees]\newline(https://www.ushmm.org/wlc/en/\newline article.php?ModuleId=10005267) in the [St. Louis](http://www.bbc.com\newline /news/magazine-27373131) exemplifies this fact. \textbackslash n\textbackslash n Great fucking point, u/*******!" \newline
        \ldots 15 more examples &
        $\textsc{Styll}_{\text{GPT-3}}$ &
        ``After going through 4 submissions that broke the peak limiter this was a great one to land on. "$~~\rightarrow$~~\newline
        ``After reviewing [four submissions](https://www.reddit.com/r/\newline gifs/comments/3a0a6j/peak\_ \newline 
        limiter\_review\_4\_submissions\newline 
        \_that/)? this one was a standout" &
        The target author heavily provides links and references in Markdown. Our method follows this stylistic pattern, but the hyperlink it generates is of course a hallucination without the model having real-world knowledge of URLs.
    \\ \hline
        ``\#HEAD ON\#\textbackslash n\textbackslash n\#APPLY DIRECTLY TO THE FOREHEAD\#" \newline
        \ldots 15 more examples &
        $\textsc{Styll}_{\text{GPT-3}}$ &
        ``Happy cake day!\textbackslash n"$~~\rightarrow$~~\newline
        ``\#BIG HAPPY BIRTHDAY\#" &
        Mimics using pairs of "\#" around a capitalized message, a very unique indicator of the target author.
    \\
    \bottomrule[\heavyrulewidth]
    \end{tabular}
    \caption{Selected example outputs generated by \textsc{Styll} with analysis.}
    \label{table:full-positive-results}
    \end{table*}
}

\newcommand{\negativeresultstable}{
    \begin{table*}[h]
    \centering
    \fontsize{9}{11}\selectfont
    \begin{tabular}{p{5.2cm}p{1.7cm}p{4.2cm}p{4.5cm}}
    \toprule[\heavyrulewidth]
    Target Style Examples & Method & Input $~~\rightarrow~~$ Output & Analysis \\ \hline
        ``CRAP DID I MISS IT" \newline
        ``HES FALLING WITH STYLE!" \newline
        ``LOL DOESNT MATTER THEY SCORED" \newline
        $\ldots$ 13 more examples &
        $\textsc{Styll}_{\text{BLOOM}}$ &
        ``McMillan sighting!"$~~\rightarrow$~~\newline
        ``McMillan got a look." &
        A low-quality paraphrase results in ``sighting'' being paraphrased to ``got a look'' which is a misinterpretation.
    \\ \hline
        ``Are they still talking about Daredevil over there?" \newline
        ``So that's what having X-ray vision kick in must be like." \newline
        ``That's more the sign on the bathroom door the next morning." \newline
        $\ldots$ 13 more examples &
        $\textsc{Styll}_{\text{GPT-3}}$ &
        ``Hey man, my wife and I lived with my mom for a bit until we got back on track. Nothing wrong with coming home for a while."$~~\rightarrow$~~\newline
        ``Hey, it's okay to come home for a little while." &
        Style transfer results in some information from the original source text being lost.
    \\ \hline
        ``Yes, yes it does." \newline
        ``Tennessee looks like trash, tbh." \newline
        ``These are two of the worst offenses in America." \newline
        $\ldots$ 13 more examples &
        $\textsc{Styll}_{\text{GPT-3}}$ &
        ``That UConn game though"$~~\rightarrow$~~\newline
        ``That UConn game was terrible." &
        Hallucination during generation results partial new information not present in the original source text. In this particular example, we suspect the hallucination may be influenced by the overall negative tone the target author uses.
    \\ \hline
        ``What is your take on gun control?" \newline
        ``I like how the maggot part is what's wrong with that statement." \newline
        ``I was never interested in Warhammer until this comment." \newline
        $\ldots$ 13 more examples &
        $\textsc{Styll}_{\text{BLOOM}}$ &
        ``Eh? Definitely bot copying Crysis."$~~\rightarrow$~~\newline
        ``What is that? What? What? I don't know what I'm talking about." &
        Hallucination during generation results in a generation that is unrecognizably different from the source text.
    \\
    \bottomrule[\heavyrulewidth]
    \end{tabular}
    \caption{Example outputs representative of the common failure modes in the generation produced by \textsc{Styll}.}
    \end{table*}
}

\newcommand{\ablationuarresultstable}{
    \begin{table*}[h]
    \centering
    \setlength{\tabcolsep}{1.8pt}
    \fontsize{9}{11}\selectfont
    \begin{tabular}{lrr|rr|rr}
    \toprule[\heavyrulewidth]
    Method ~~~ & \multicolumn{2}{c|}{Random}& \multicolumn{2}{c|}{Single}& \multicolumn{2}{c}{Diverse}\\
           & \multicolumn{1}{c}{\textsc{Confusion}} & \multicolumn{1}{c|}{\textsc{Joint}} & \multicolumn{1}{c}{\textsc{Confusion}} & \multicolumn{1}{c|}{\textsc{Joint}} & \multicolumn{1}{c}{\textsc{Confusion}} & \multicolumn{1}{c}{\textsc{Joint}} \\ \hline
    - \textsc{Des}  & 0.00 & \textbf{0.18} & 0.00 & \textbf{0.33} & 0.22 & \textbf{0.44} \\
    + \textsc{Des}  & \underline{0.11} & 0.10 & \underline{0.33} & 0.29 & \underline{0.33} & 0.38 \\
    \bottomrule[\heavyrulewidth]
    \end{tabular}
    \caption{Automatic evaluation metrics over smaller dataset variants of 3 source authors and 3 target authors for our method \textsc{Styll} ablating the use of style descriptors in the UAR Embedding space. \textsc{Des} indicates usage of style descriptors in the few-shot style transfer prompt described in step 3 of Section 4.}
    \end{table*}
}

\newcommand{\strapvsstylltable}{
    \begin{table*}[h]
    \centering
    \fontsize{9}{11}\selectfont
    \begin{tabular}{p{3.5cm}p{5cm}p{4.5cm}}
    \toprule[\heavyrulewidth]
    Target Style Examples & Input $~~\rightarrow~~$ \textsc{Strap} Output & Input $~~\rightarrow~~$ \textsc{Styll} Output \\ \hline
        ``Who wants salad for lunch?" \newline
        ``Dirt road?  Extra tire squeal for you!" \newline
        ``Divorce can ruin a person financially despite their steady income." \newline
        \ldots 13 more examples &
        ``Who are pretty much ``buying" the development from Aalto."$~~\rightarrow$~~\newline
        ``who's buying A who's buying Aal who's buying Aalto's development pretty much" &
        ``Who are pretty much ``buying" the development from Aalto."$~~\rightarrow$~~\newline
        ``Aalto is getting financial support from a variety of sources in order to develop their project."

    \\ \hline
        ``If you take my hat, I'm taking some of your teeth" \newline
        ``If you would die tomorrow, what do you want to eat?"\newline
        ``Be. The ball." \newline
        \ldots 13 more examples &
        ``Lots of misconceptions about airbag deployment. It's not necessarily the point of impact that determines whether the bags deploy or not, it is the deceleration of the occupants as determined by the airbag system. Caravans are relatively soft so the impact was likely not severe enough."$~~\rightarrow$~~\newline
        ``The airbag is notthe airbag .The airbag is not the airbag is not the airbag is not the airbag is not the airbag is not the airbag" &
        ``Lots of misconceptions about airbag deployment. It's not necessarily the point of impact that determines whether the bags deploy or not, it is the deceleration of the occupants as determined by the airbag system. Caravans are relatively soft so the impact was likely not severe enough."$~~\rightarrow$~~\newline
        ``There are a lot of myths about airbag deployment. One of them is that it is always based on the impact. The truth is that it is based on the deceleration of the occupants. Caravans are not as stiff as other types of vehicles, so the impact may have been less severe." \newline
    \\ \hline
        ``anyone have the full source?" \newline
        ``You're a idiot."\newline
        ``No beta? No way." \newline
        \ldots 13 more examples &
        ``Nice to see Madoka there enjoying it with her."$~~\rightarrow$~~\newline
        ``. . The first thing is the first thing is the first thing is the first"
        &
        ``Nice to see Madoka there enjoying it with her."$~~\rightarrow$~~\newline
        ``It's refreshing to see Madoka having fun."
    \\
    \bottomrule[\heavyrulewidth]
    \end{tabular}
    \caption{Example outputs generated by $\textsc{Strap}_{p=0.0}$  and $\textsc{Styll}_{\text{GPT-3}}$ on the ``Diverse'' dataset. Examples are chosen at random as to be a representative sample of general output quality. Not only does \textsc{Styll} outperform \textsc{Strap} on the \textsc{Joint} automatic evaluation metric as discussed in Section 6, in our low-resource setting we qualitatively observe that its generated outputs are of far higher quality and fluency. \textsc{Strap} greatly suffers with issues of generation repetition and coherence when trained with such little training data.}
    \end{table*}
}

\newcommand{\llmuarresultstable}{
    \begin{table*}[h]
    \centering
    \setlength{\tabcolsep}{1.8pt}
    \fontsize{9}{11}\selectfont
    \begin{tabular}{lrrrr|rrrr|rrrr}
    \toprule[\heavyrulewidth]
    Method ~~~ & \multicolumn{4}{c|}{Random}& \multicolumn{4}{c|}{Single}& \multicolumn{4}{c}{Diverse}\\
           & \multicolumn{1}{c}{\textsc{Away}} & \multicolumn{1}{c}{\textsc{Towards}}& \multicolumn{1}{c}{\textsc{Sim}} & \multicolumn{1}{c|}{\textsc{Joint}} & \multicolumn{1}{c}{\textsc{Away}} & \multicolumn{1}{c}{\textsc{Towards}}& \multicolumn{1}{c}{\textsc{Sim}} & \multicolumn{1}{c|}{\textsc{Joint}} & \multicolumn{1}{c}{\textsc{Away}} & \multicolumn{1}{c}{\textsc{Towards}}& \multicolumn{1}{c}{\textsc{Sim}} & \multicolumn{1}{c}{\textsc{Joint}} \\ \hline
    $\text{GPT-2}_{\text{1.5B}}$  & 0.92 & 0.17 & 0.10 & 0.17 & 0.97 & 0.13 & 0.13 & 0.16 & 0.93 & 0.14 & 0.14 & 0.17 \\
    $\text{GPT-3}_{\text{6.7B}}$  & 0.78 & 0.07 & 0.45 & 0.23 & 0.91 & 0.11 & 0.48 & 0.29 & 0.87 & 0.12 & 0.44 & 0.30 \\
    $\text{GPT-J}_{\text{6B}}$  & 0.71 & 0.11 & 0.56 & \textbf{0.35} & 0.87 & 0.15 & 0.60 & 0.39 & 0.77 & 0.12 & 0.60 & \textbf{0.36} \\
    $\text{OPT}_{\text{6.7B}}$  & 0.74 & 0.13 & 0.43 & 0.31 & 0.89 & 0.14 & 0.44 & 0.32 & 0.82 & 0.13 & 0.50 & 0.35 \\
    $\text{BLOOM}_{\text{7.1B}}$  & 0.70 & 0.11 & 0.54 & 0.34 & 0.86 & 0.16 & 0.57 & \textbf{0.40} & 0.76 & 0.12 & 0.58 & \textbf{0.36} \\
    $\text{FLAN-T5}_{\text{3B}}$  & 0.69 & 0.05 & 0.61 & 0.21 & 0.88 & 0.06 & 0.62 & 0.22 & 0.77 & 0.06 & 0.63 & 0.26 \\
    \bottomrule[\heavyrulewidth]
    \end{tabular}
    \caption{Automatic evaluation metrics for our method \textsc{Styll} applied with different large language models for step 3 of Section 4 over three dataset variants in the UAR Embedding space.}
    \end{table*}
}

\newcommand{\llmstyleresultstable}{
    \begin{table*}[h]
    \centering
    \setlength{\tabcolsep}{1.8pt}
    \fontsize{9}{11}\selectfont
    \begin{tabular}{lrrrr|rrrr|rrrr}
    \toprule[\heavyrulewidth]
    Method ~~~ & \multicolumn{4}{c|}{Random}& \multicolumn{4}{c|}{Single}& \multicolumn{4}{c}{Diverse}\\
           & \multicolumn{1}{c}{\textsc{Away}} & \multicolumn{1}{c}{\textsc{Towards}}& \multicolumn{1}{c}{\textsc{Sim}} & \multicolumn{1}{c|}{\textsc{Joint}} & \multicolumn{1}{c}{\textsc{Away}} & \multicolumn{1}{c}{\textsc{Towards}}& \multicolumn{1}{c}{\textsc{Sim}} & \multicolumn{1}{c|}{\textsc{Joint}} & \multicolumn{1}{c}{\textsc{Away}} & \multicolumn{1}{c}{\textsc{Towards}}& \multicolumn{1}{c}{\textsc{Sim}} & \multicolumn{1}{c}{\textsc{Joint}} \\ \hline
    $\text{GPT-2}_{\text{1.5B}}$  & 0.91 & 0.20 & 0.10 & 0.13 & 0.92 & 0.25 & 0.13 & 0.15 & 0.93 & 0.18 & 0.14 & 0.14 \\
    $\text{GPT-3}_{\text{6.7B}}$  & 0.82 & 0.13 & 0.45 & 0.22 & 0.92 & 0.31 & 0.48 & 0.39 & 0.93 & 0.23 & 0.44 & 0.33 \\
    $\text{GPT-J}_{\text{6B}}$  & 0.77 & 0.17 & 0.56 & 0.31 & 0.87 & 0.29 & 0.60 & \textbf{0.43} & 0.86 & 0.24 & 0.60 & \textbf{0.41} \\
    $\text{OPT}_{\text{6.7B}}$  & 0.79 & 0.19 & 0.43 & 0.29 & 0.90 & 0.28 & 0.44 & 0.34 & 0.88 & 0.25 & 0.50 & 0.38 \\
    $\text{BLOOM}_{\text{7.1B}}$  & 0.74 & 0.19 & 0.54 & \textbf{0.32} & 0.84 & 0.30 & 0.57 & 0.40 & 0.83 & 0.26 & 0.58 & \textbf{0.41} \\
    $\text{FLAN-T5}_{\text{3B}}$  & 0.81 & 0.08 & 0.61 & 0.19 & 0.91 & 0.14 & 0.62 & 0.25 & 0.87 & 0.10 & 0.63 & 0.22 \\
    \bottomrule[\heavyrulewidth]
    \end{tabular}
    \caption{Automatic evaluation metrics for our method \textsc{Styll} applied with different large language models for step 3 of Section 4 over three dataset variants in the Style Embedding space.}
    \end{table*}
}

\newcommand{\decodinguarresultstable}{
    \begin{table*}[h]
    \centering
    \setlength{\tabcolsep}{1.8pt}
    \fontsize{9}{11}\selectfont
    \begin{tabular}{lrrrr|rrrr|rrrr}
    \toprule[\heavyrulewidth]
    Method ~~~ & \multicolumn{4}{c|}{Random}& \multicolumn{4}{c|}{Single}& \multicolumn{4}{c}{Diverse}\\
           & \multicolumn{1}{c}{\textsc{Away}} & \multicolumn{1}{c}{\textsc{Towards}} & \multicolumn{1}{c}{\textsc{Sim}} & \multicolumn{1}{c|}{\textsc{Joint}} & \multicolumn{1}{c}{\textsc{Away}} & \multicolumn{1}{c}{\textsc{Towards}} & \multicolumn{1}{c}{\textsc{Sim}} & \multicolumn{1}{c|}{\textsc{Joint}} & \multicolumn{1}{c}{\textsc{Away}} & \multicolumn{1}{c}{\textsc{Towards}} & \multicolumn{1}{c}{\textsc{Sim}} & \multicolumn{1}{c}{\textsc{Joint}} \\ \hline
    $t = 0.1$  & 0.67 & 0.06 & 0.65 & 0.24 & 0.85 & 0.11 & 0.66 & 0.32 & 0.75 & 0.08 & 0.68 & 0.31 \\
    $t = 0.7$  & 0.70 & 0.11 & 0.54 & 0.34 & 0.86 & 0.16 & 0.57 & \textbf{0.40} & 0.76 & 0.12 & 0.58 & \textbf{0.36} \\
    $t = 0.9$  & 0.73 & 0.15 & 0.42 & \textbf{0.35} & 0.87 & 0.18 & 0.47 & 0.38 & 0.79 & 0.15 & 0.46 & 0.35 \\
    \bottomrule[\heavyrulewidth]
    \end{tabular}
    \caption{Automatic evaluation metrics for our method \textsc{Styll} applied with different settings of \texttt{temperature} for step 3 of Section 4 over three dataset variants in the UAR Embedding space.}
    \end{table*}
}

\begin{abstract}
Authorship style transfer involves altering text to match the style of a target author whilst preserving the original meaning. Existing unsupervised approaches like \textsc{Strap} have largely focused on style transfer to target authors with many examples of their writing style in books, speeches, or other published works. This high-resource training data requirement (often greater than 100,000 words) makes these approaches primarily useful for style transfer to published authors, politicians, or other well-known figures and authorship styles, while style transfer to non-famous authors has not been well-studied. We introduce the \textit{low-resource authorship style transfer} task, a more challenging class of authorship style transfer where only a limited amount of text in the target author's style may exist. In our experiments, we specifically choose source and target authors from Reddit and style transfer their Reddit posts, limiting ourselves to just 16 posts (on average $\approx$~500 words) of the target author's style. Style transfer accuracy is typically measured by how often a classifier or human judge will classify an output as written by the target author. Recent authorship representations models excel at authorship identification even with just a few writing samples, making automatic evaluation of this task possible for the first time through evaluation metrics we propose. Our results establish an in-context learning technique we develop as the strongest baseline, though we find current approaches do not yet achieve mastery of this challenging task. We release our data and implementations to encourage further investigation.
\end{abstract}

\section{Introduction}
\label{sec:introduction}

\textit{Authorship style transfer} involves applying the style of some target author's texts to a text of a source author \citep{tstsurvey}. \textit{Style}, in this context, has typically included, but not been limited to, linguistic attributes such as syntax, grammar, spelling, lexical choice, and punctuation choice \citep{styleemb}. After performing style transfer, the output should closely match the desired target author's style while still preserving the meaning of the original text \citep{strap}.

\begin{figure}[t]
    \centering
    \includegraphics[width=235px]{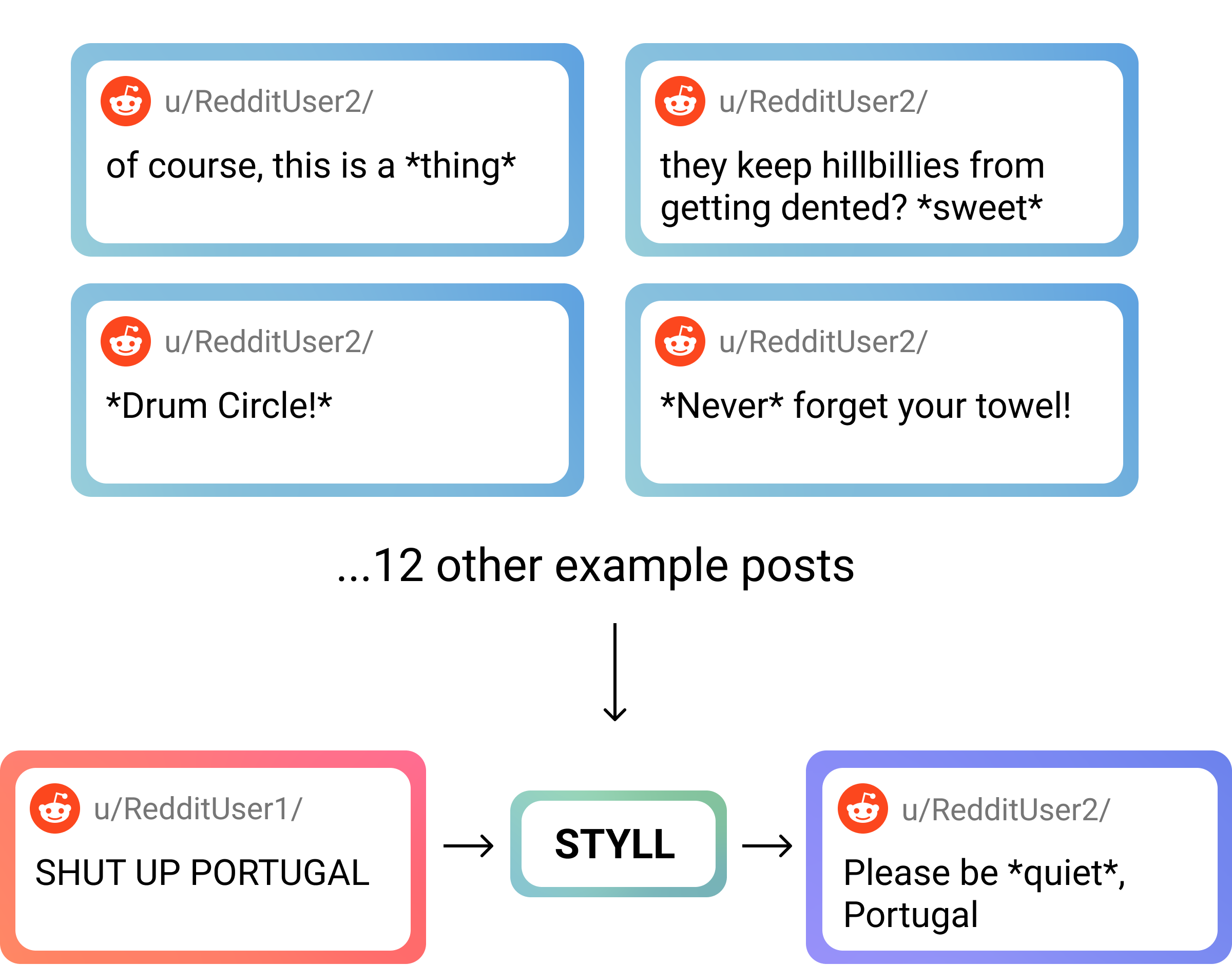}
    \caption{An actual output of \textsc{Styll} on the unsupervised low-resource authorship style transfer task between two Reddit users using just 16 Reddit posts as examples of the target style.}
    \label{fig:styll-figure}
\end{figure}

Prior work in authorship style transfer has largely been limited to the domain of style transfer where many examples of the target author's style exists \citep{shakespeare,bible,strap}. For example, style transfer to the style of William Shakespeare utilizing a large collection of his published works as examples of the target style. We will define this type of authorship style transfer as \textit{high-resource authorship style transfer}, taking inspiration from terminology used in machine translation (MT) to denote data requirements. In MT, distinct techniques and models are often used to perform translation for low-resource languages, where very little example text of that language may exist and therefore models and techniques used for translation of high-resource languages may not be effective \citep{lrmt}.

\begin{figure}[t]
    \centering
    \includegraphics[width=225px]{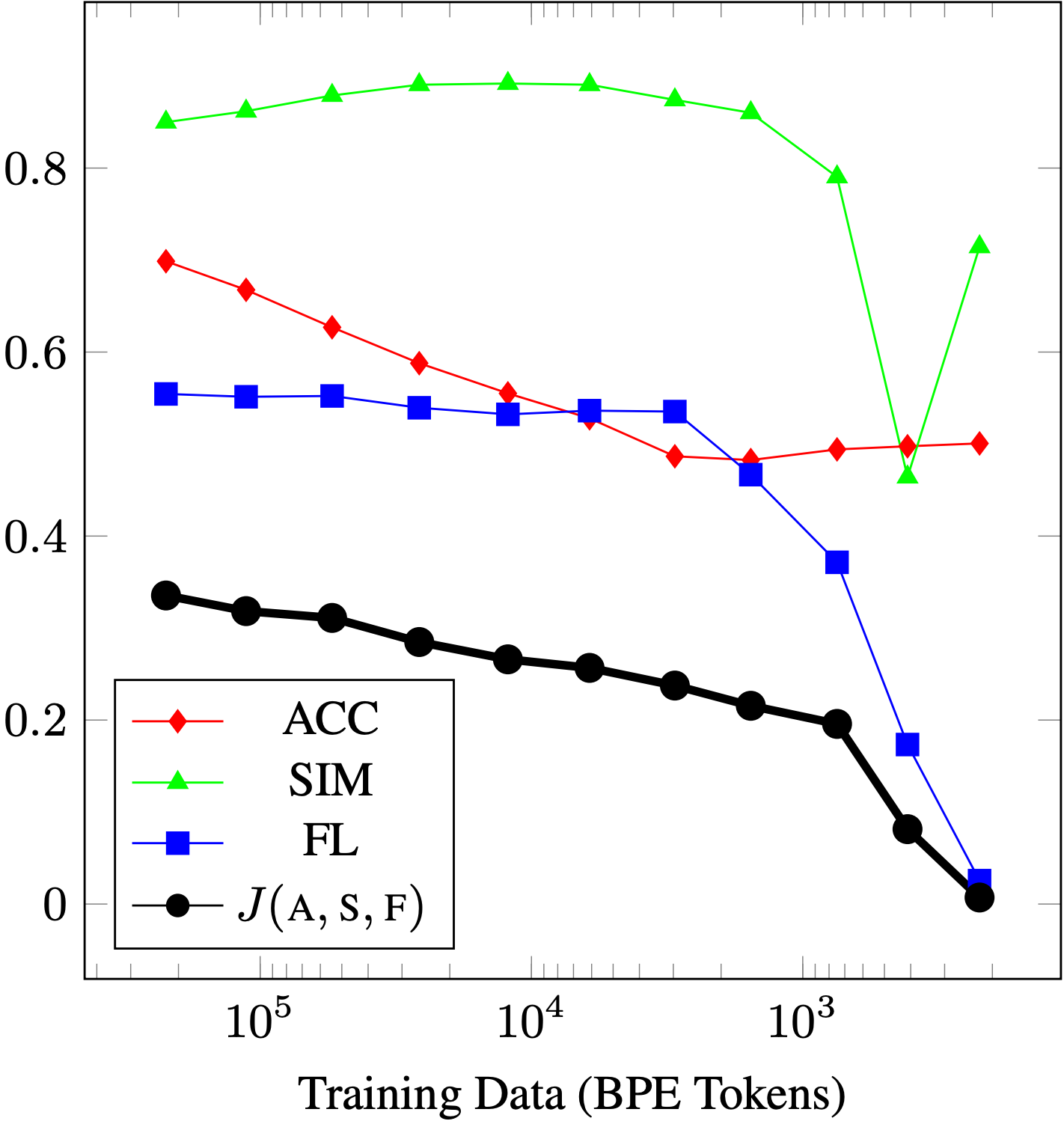}
    \caption{Scores of various evaluation metrics and a joint score, $J(\textsc{a}, \textsc{s}, \textsc{f})$, on style transfer outputs produced by $\textsc{Strap}_{p=0.0}$ \citep{strap} on the Shakespeare author imitation dataset \citep{shakespeare} given decreasing amounts of training example tokens. \textsc{Strap}'s performance falls off as the number of training tokens decreases and drops precipitously in the low-resource setting.}
    \label{fig:strap-accuracy-figure}
\end{figure}

We define \textit{low-resource authorship style transfer} as style transfer to target authors who only have a limited amount of example text. In this paper, we limit ourselves to using just 16 social media posts or on average $\approx$~500 byte-pair encoded (BPE) tokens of a target author as examples of their style. In Figure \ref{fig:styll-figure}, we visualize the setup of our task. The ability to style transfer to less well-known figures and average users is motivated by several downstream applications, such as imitating the writing style of an average user in commercial text-editing software, customizing the style of conversational interfaces to users, or even, in the case of performing style transfer over programming language text, customizing generated code in the style of a particular programmer or existing codebase. Outside of commercial applications, research on this task can have positive broader impacts. Recent work by \citet{luar} demonstrates an authorship identification (AID) model that with just 16 Reddit posts identifies the author from among hundreds of thousands of candidates with high accuracy. Research towards low-resource authorship style transfer can help provide a utility for authorship obfuscation adversarial to AID. For targets of malevolent actors using such AID systems, like political dissidents, low-resource style transfer can act as a privacy preserving shield analogous to other anonymizing tools such as VPN networks. It is important to note that high-resource authorship style transfer methods often utilize techniques that are not performant in the low-resource setting. For example, the \textsc{Strap} method involves fine-tuning a GPT-2 model, requiring both a train and validation dataset with examples \citep{gpt2,strap}. Fine-tuning GPT-2 successfully would require significantly more than the $\approx$~500 BPE tokens found in 16 Reddit post examples. In \citet{strap}, the smallest train set used (``poetry'') had $\approx$~290K BPE tokens, while the largest (``tweets'') had $\approx$~67.7M BPE tokens. In Figure \ref{fig:strap-accuracy-figure}, we demonstrate the fall off of \textsc{Strap}'s performance as the amount of example text in the target author's style decreases.

With this introduction, we summarize the primary contributions of this work as follows:

\begin{enumerate}
    \item We define and motivate the low-resource authorship style transfer task.
    \item We draw three dataset variants from a dataset of Reddit authors \citep{accountlinking} to evaluate the task under different content scenarios.
    \item We propose a method and metrics for automatic evaluation of the task utilizing authorship and style representation embeddings and conduct a human evaluation.
    \item We develop a comprehensive set of baselines for the low-resource authorship style transfer task. We establish \textsc{Styll} (\textbf{Sty}le Transfer with \textbf{L}arge \textbf{L}anguage Models), an in-context learning technique we propose, as the strongest baseline we evaluate for this task.
\end{enumerate}

\section{Related Work}
\label{sec:related-work} 

\textit{Style attribute transfer} is an adjacent task that involves transforming text on a single particular style dimension (e.g. ``relaxed" $\rightarrow$ ``annoyed") \citep{lample2018multiple,Lample2019MultipleAttributeTR,gst}. Few-shot style transfer approaches have been primarily demonstrated for the style attribute transfer task. \citep{vae,styletransferrecipe}. \citet{textsettr} introduces an approach leveraging style vector conditioned language models to demonstrate few-shot style attribute transfer, but also experiments with few-shot authorship style transfer (100 examples) to the style of Shakespeare, a high-resource author. Given they leverage pre-trained language models, their few-shot setting benefits from the many examples of Shakespeare's famous style indirectly seen during self-supervised pre-training, making that experiment setup distinct from the low-resource non-famous authors we target in this work.

\section{Dataset}
\label{sec:dataset}

To perform low-resource authorship style transfer, we first choose a realistic domain to select source and target authors of the style transfer. We choose authors from a Reddit corpus \citep{accountlinking} as our source and target users. These users are not likely to be well-published figures, but rather average anonymous social media users. Each Reddit user has 16 posts, so we style transfer 16 posts of a source author to the target author's style using the target author's 16 posts as examples of their style. Since we have no parallel examples between the source and target author to supervise learning, the setting here is \textit{unsupervised} low-resource authorship style transfer.

For the authorship style transfer task, style and content can be deeply interwoven and it can sometimes be impossible to independently separate the two \citep{tstsurvey}. For example, it may not be possible to plausibly or convincingly style transfer text written by a video game enthusiast source author to the style of a target author who is a distinguished politician. \citet{tstsurvey} recommends limiting the task to ``scenarios where the attribute and semantics can be approximately separated.'' For this reason, we draw three distinct dataset variants from Reddit that help test the effectiveness low-resource authorship style transfer can achieve in different scenarios. In each variant, we choose 15 source authors and 15 target authors and perform style transfer over all pairs, resulting in 225 unique style transfer source-target author pairs per variant. Each source-target author pair has 16 Reddit posts, so we style transfer 3,600 posts per dataset variant.

\begin{itemize}
    \item \textbf{Random}: 15 source authors and 15 target authors chosen at random.
    \item \textbf{Single}: 15 source authors and 15 target authors whose 16 posts all belong to the most common subreddit, ``r/CFB'' (a subreddit about college football). In this variant, using subreddit metadata, we attempt to control for content by ensuring all text involves discussing a consistent topic.
    \item \textbf{Diverse}: 15 source authors and 15 target authors whose 16 posts belong to at least 13 different subreddits. In this variant, using subreddit metadata, we attempt to ensure any source author or target author posts about a diverse number of topics.
\end{itemize}

We hypothesize authorship style transfer over the ``Random'' dataset may be difficult or impossible due to the nature of the stark differences in topics source and target authors regularly discuss.

\section{Method}
\label{sec:method}

Since low-resource authorship style transfer has not been widely researched, we develop a comprehensive set of baselines to evaluate, including manual handcrafted baselines to perform style transfer in addition to evaluating against a state-of-the-art open source technique for unsupervised style transfer, \textsc{Strap}\footnote{We evaluate \textsc{Strap} over three values of $p$ during generation: 0.0, 0.6, 0.9.} \citep{strap}. Minor implementation details can be found in the Appendix.

\paragraph{$\textsc{Copy}_{\textsc{src}}$} A na\"ive baseline that simply copies the source author's post without modifying it at all.
\paragraph{$\textsc{Copy}_{\textsc{tgt}}$} A na\"ive baseline that simply copies a target author's post without modifying it at all.
\paragraph{\textsc{Capi}} Computes the probability distribution of the target author's posts being one of three capitalization styles: 1) ``uppercase'', 2) ``lowercase'', and 3) ``sentence case''. The capitalization style of the source author's posts is then transformed following the probability distribution.
\paragraph{\textsc{Cont}} Computes the probability distribution of the target author's posts using or not using contractions. The contraction style of the source author's posts are then transformed following the probability distribution.
\paragraph{\textsc{Synm}} Swaps words\footnote{\label{wordswap}Swapped words are transformed to match the inflection and case of the original word with the package \texttt{lemminflect}.} from the source author's posts with a word from the target author's posts when the words are synonyms according to WordNet's synsets.
\paragraph{\textsc{Punc}} Each of the source author's posts is transformed to swap the end mark punctuation of sentences with the the end mark punctuation used in the target author's posts.
\paragraph{\textsc{Emoj}} Highly distinctive non-ASCII strings or strings of at least two punctuation characters that are not end marks that are found in the target author's posts are injected into the source author's posts with the same frequency.
\paragraph{\textsc{Para}} To determine if step 3 of our procedure in Section \ref{sec:styll} is actually effective, we evaluate only paraphrasing\footnote{$\textsc{Para}_{\textsc{Neu}}$ refers to using $\text{GPT-3}_{\text{6.7B}}$ for step 1, while $\textsc{Para}_{\textsc{Div}}$ refers to using the diverse paraphrase model from \citet{strap} for step 1.} the source author's posts as a baseline since paraphrasing alone should move the style away from the source author's style (but may not help move towards the target author's style).
\paragraph{\textsc{Ling}}
\label{para:ling}
Composes all of the prior targeted linguistic baselines (\textsc{Capi}, \textsc{Cont}, \textsc{Synm}, \textsc{Punc}, and  \textsc{Emoj}) together. This baseline represents a robust handcrafted solution to low-resource authorship style transfer. While we expect this baseline to perform reasonably well, this technique is not trainable. Therefore, it leaves little room for future improvement, where as we can see that \textsc{Styll}, a baseline we propose in the next section, improves with model scale in Appendix B.
\paragraph{\textsc{Bert}} Swaps words\footnotemark[3] or punctuation from the source author's posts with a word or punctuation from the target author's posts when the cosine similarity between the average BERT embeddings \citep{bert,roberta} of the two tokens is $\geq$0.6  and the part-of-speech tags match, following \citet{topgunn}.

\subsection{Style Transfer with Large Language Models (\textsc{Styll})}
\label{sec:styll}

We also develop a new unsupervised authorship style transfer procedure using few-shot prompting, a baseline we call \textsc{Styll} (\textbf{Sty}le Transfer with \textbf{L}arge \textbf{L}anguage Models). It is well-established that large language models (LLMs) are strong zero-shot and few-shot performers and require very little to no data to perform previously unseen tasks by utilizing in-context learning (ICL) \citep{gpt3}. ICL has been applied to other low-resource tasks successfully such as low-resource MT \citep{bidi} and has even been used to perform style attribute transfer \citep{styletransferrecipe}. These properties indicate promising potential on our challenging low-resource task.  One practical benefit of this approach is that our technique is simple and requires no fine-tuning and can be immediately used on new authors where as \textsc{Strap} requires fine-tuning a model per authorship style. Although LLMs have anecdotally been observed to perform style transfer, for example to common authors such as Shakespeare, to our knowledge we are the first to systematically evaluate their ability to perform in-context learning of arbitrary and diverse authorship styles.

To perform unsupervised style transfer without any parallel data between source and target author styles, we follow \citet{strap} and reformulate style transfer as a paraphrase and inverse paraphrase task. We paraphrase the target author's example posts to a ``neutral'' style. With this, we can build a synthetic dataset with parallel examples from the ``neutral'' style to the target author's style. These synthetic parallel examples are then used with ICL to style transfer a source author's post, also paraphrased in the ``neutral'' style, to the target author's style. Our method consists of three steps:

\begin{itemize}
\item[] \textbf{Step 1}: Source author posts and target author posts are paraphrased to a ``neutral'' style using a zero-shot prompt:

\begin{displayquote}
{
Passage: \textbf{\textcolor{black}{[\textit{Post to Paraphrase}]}}
\\\\
Paraphrase the passage in a simple neutral style.
\\\\
Rewrite:
}
\end{displayquote}

For example, the post ``\textit{Eh, that Nissa looks pretty competitive.}'' by one of the Reddit authors is neutrally paraphrased to ``\textit{Nissa seems to be a very competitive person.}'' by this prompt.

\item[] \textbf{Step 2}: The style of each target author's example posts are described in a few comma-separated adjectives we call \textit{style descriptors} using a zero-shot prompt:
\begin{displayquote}
{
Passage: \textbf{\textcolor{black}{[\textit{Target Author Example Post \#1}]}}\\
Passage: \textbf{\textcolor{black}{[\textit{Target Author Example Post \#2}]}}\\
\ldots \\
Passage: \textbf{\textcolor{black}{[\textit{Target Author Example Post \#3}]}}\\
List some adjectives, comma-separated, that describe the writing style of the author of these passages:
}
\end{displayquote}

The style descriptors ``\textit{clear, concise, persuasive, intelligent}'' are an example generation from this prompt for one of the Reddit authors.

\item[] \textbf{Step 3}: Source author posts undergo style transfer to the target author's style using a few-shot prompt following the prompt template used in \citet{styletransferrecipe}:

\begin{displayquote}
{
Here is some text: \{\textbf{\textcolor{black}{[\textit{Neutral Paraphrase of Target Example \#1}]}}\} Here is a rewrite of the text that is more \textbf{\textcolor{black}{[\textit{Style Descriptors}]}}: \{\textbf{\textcolor{black}{[\textit{Target Example \#1}]}}\} Here is some text: \{\textbf{\textcolor{black}{[\textit{Neutral Paraphrase of Target Author Example Post \#2}]}}\} Here is a rewrite of the text that is more \textbf{\textcolor{black}{[\textit{Style Descriptors}]}}: \{\textbf{\textcolor{black}{[\textit{Target Author Example Post \#2}]}}\}\ldots Here is some text: \{\textbf{\textcolor{black}{[\textit{Neutral Paraphrase of Target Example \#16}]}}\} Here is a rewrite of the text that is more \textbf{\textcolor{black}{[\textit{Style Descriptors}]}}: \{\textbf{\textcolor{black}{[\textit{Target Example \#16}]}}\} Here is some text: \{\textbf{\textcolor{black}{[\textit{Neutral Paraphrase of Source Author Post}]}}\} Here is a rewrite of the text that is more \textbf{\textcolor{black}{[\textit{Style Descriptors}]}}: \{
}
\end{displayquote}

\end{itemize}

Steps 1 and 2 are preprocessing steps, while step 3 is where style transfer is performed using the output of steps 1 and 2. The use of an intermediate output from an LLM in a subsequent prompt is inspired by the chain-of-thought and self-ask prompting techniques \citep{chainofthought,selfask}. For step 3, to demonstrate \textsc{Styll}'s generalizability across models, we evaluate the performance of $\text{GPT-2}_{\text{1.5B}}$ \citep{gpt2}, $\text{GPT-3}_{\text{6.7B}}$, $\text{GPT-J}_{\text{6B}}$ \citep{gptj},  $\text{OPT}_{\text{6.7B}}$ \citep{opt}, $\text{BLOOM}_{\text{7.1B}}$ \citep{bloom}, and $\text{FLAN-T5}_{\text{3B}}$ \citep{flant5} in Appendix B. In this paper, we show results using $\text{GPT-3}_{\text{6.7B}}$ and the open source $\text{BLOOM}_{\text{7.1B}}$ model for step 3. When using $\text{GPT-3}_{\text{6.7B}}$ for step 3, we use $\text{GPT-3}_{\text{6.7B}}$ for steps 1 and 2. When using an open source model for step 3, we use the diverse paraphrase model from \citet{strap} for step 1 and $\text{FLAN-T5}_{\text{3B}}$ for step 2 to maintain a procedure that is open source and reproducible. We ablate the use of style descriptors in the few-shot prompt in step 3. The results of this ablation can be found in Appendix D.

\postiveresultstable

\section{Evaluation}
\label{sec:evaluation}

Automatic evaluation metrics for style transfer proposed by \citet{strap} and others \citep{styleattributetransfer,tstsurvey,styletransferrecipe} typically consider: 1) accuracy of the style transfer, 2) meaning preservation, and sometimes, 3) fluency of the output. We follow this existing framework to propose automatic evaluation metrics for low-resource authorship style transfer.

To measure accuracy of the style transfer, prior work trained classifiers to perform authorship or style identification of text. If the style transfer output could manipulate the classifier's decision, it could be considered successful. In our low-resource setting, where we only have 16 posts per author, it would be improbable we could attain an accurate classifier this way. Instead, we propose utilizing authorship representation embeddings or style representation embeddings to measure the accuracy of the transfer. These embeddings create a singular vector that represents the authorship or the style of a set of texts in a continuous vector space. By measuring movement away from the source author and movement towards the target author in this space, we can achieve an automatic measure of style transfer accuracy\footnote{We want to explicitly measure movement both away from the source author and towards the target author since, in a vector space, it is possible to have multiple locations equidistant to the target author with some being closer to the source author than others.}. We evaluate our metrics over two embedding spaces, the Universal Author Representation (UAR) Embeddings \citep{luar}, which capture style and content to represent authorship with a continuous vector representation, and Style Embeddings, which aim to capture only style in a continuous vector representation \citep{styleemb}, but are trained on far less data than the UAR Embeddings. We primarily utilize UAR Embeddings in this work, but provide results for Style Embeddings in Appendix F for reference.

For meaning preservation, we use the Mutual Implication Score \citep{mis} between the output and the original text. We omit measuring fluency of the output as our Reddit posts do not consistently register as fluent\footnote{For example, on the ``Single" dataset, the target author texts themselves range widely in fluency scores from 0.57 to 0.92 with a standard deviation of 0.06.} due to high usage of slang, jargon, and informal syntax and grammar. This style of informal text is not well-represented in the CoLA corpus \citep{cola}, which the fluency model used by \citet{strap} is trained on. For this reason, fluency ratings would add noise to our joint metric. Selected generations can be found in Table \ref{table:positive-results} and random generations representative of the general output quality appear in Appendix G.
 
For notation purposes, we represent the set of source authors as $S$ and the set of target authors as $T$. We represent the set of any given author $a$'s 16 posts with $P_{a}$ and source author $s$'s 16 posts style transferred to the target author $t$'s style with $P_{s \rightarrow t}$. We let $\vec{R}(P)$ represent a single UAR Embedding produced over a set of posts $P$. We use $\text{MIS}(P_{a},~ P_{b})$ to denote the average Mutual Implication Score between two sets of posts by authors $a$ and $b$. Finally, we use $\mathcal{S}(\vec{u}, \vec{v})$ to refer to the similarity measure\footnote{$\text{sim}(\vec{u}, \vec{v}) = \left(1-\arccos \left(\frac{\vec{u} \cdot \vec{v}}{\|\vec{u}\|\|\vec{v}\|}\right) / \pi\right)$} found in \citet{angularsim} between $\vec{u}$ and $\vec{v}$ scaled to 0 and 1, that is $\mathcal{S}(\vec{u}, \vec{v}) = \frac{\text{sim}(\vec{u}, \vec{v}) + 1}{2}$ and we define the complement $\mathcal{S}_{c}(\vec{u}, \vec{v}) = 1 - \mathcal{S}(\vec{u}, \vec{v})$.

\paragraph{$\textsc{Away}(s, t)$}
The away score measures how far the style transferred posts are from the source author's posts in the representational embedding space as a percentage of how far the target author's posts are from the source author's posts (the ideal distance).
\[
\frac{\text{min}\Bigl(\mathcal{S}_{c}\bigl(\vec{R}(P_{s \rightarrow t}),~\vec{R}(P_{s})\bigr),~\mathcal{S}_{c}\bigl(\vec{R}(P_{t}),~ \vec{R}(P_{s})\bigr)\Bigr)}{\mathcal{S}_{c}\bigl(\vec{R}(P_{t}),~ \vec{R}(P_{s})\bigr)}
\]

\uarembeddingresultstable
\paragraph{$\textsc{Towards}(s, t)$}
The towards score measures how far towards the target author's  posts the style transferred posts moved in the representational embedding space as a percentage of the maximum possible distance they could move towards the target author's posts.
\[
\frac{\text{max}\Bigl(\mathcal{S}\bigl(\vec{R}(P_{s \rightarrow t}),~\vec{R}(P_{t})\bigr) - \mathcal{S}\bigl(\vec{R}(P_{s}),~ \vec{R}(P_{t})\bigr),~0\Bigr)}{\mathcal{S}_{c}\bigl(\vec{R}(P_{s}),~ \vec{R}(P_{t})\bigr)}
\]

\paragraph{$\textsc{Sim}(s, t)$}
To measure meaning preservation we compute the change in average MIS between the style transferred posts and the source author's posts from the average MIS between the target author's posts and the source author's posts as a percentage of the maximum change possible.
\[
\frac{\text{max}\Bigl(\text{MIS}(P_{s \rightarrow t},~P_{s}) - \text{MIS}(P_{t},~ P_{s}),~0\Bigr)}{1 - \text{MIS}(P_{t},~ P_{s})}
\]
\paragraph{$\textsc{Joint}(s, t)$}
\label{para:joint}
The joint score composes a single balanced evaluation metric from other metrics computed with the geometric mean $\mathcal{G}$, following prior work \citep{strap}. Style transfer accuracy and meaning preservation is weighed equally; in practice, however, we may prefer to sacrifice some meaning preservation for higher style transfer accuracy.
For this reason, and since style and content can often be interwoven, this evaluation metric is useful as a quantitative measure, but is not a definitive measure of style transfer quality, similar to metrics like BLEU used in MT \citep{bleurole}.
\[
\mathcal{G}\Bigl(\bigl[\mathcal{G}\bigl([\textsc{Away}(s, t), \textsc{Towards}(s, t)]\bigr), \textsc{Sim}(s, t)\bigr]\Bigr)
\]

\section{Results}
\label{sec:results}
The results of our automatic evaluation metrics can be found in Table \ref{table:uar-embedding-results} and we find \textsc{Styll} to be the strongest baseline for this task. Example outputs of \textsc{Styll} and analysis can be found in Table \ref{table:positive-results}. We observe common failure modes to include standard generation errors produced by LLMs such as hallucinations as well as imperfect paraphrases leading to undesirable phrasing; annotated examples with analysis can be found in Appendix I.

\paragraph{Automatic Evaluation} On the \textsc{Joint} metric, which accounts for style transfer accuracy and meaning preservation, our method outperforms the comprehensive set of baselines as well as \textsc{Strap} on the ``Single'' and ``Diverse'' dataset variants, which help control for content or ensure authors discuss diverse topics, increasing the likelihood that authorship style transfer is plausible between any given source-target author pair. On the ``Random'' variant, where authorship style transfer may be undefined or impossible, we find \textsc{Styll} performs closer, but still slightly outperforms baselines. Appendix G demonstrates the low-quality nature of \textsc{Strap}'s outputs in the low-resource setting compared to \textsc{Styll} as predicted by Figure \ref{fig:strap-accuracy-figure}.

\uarretrievalresultstable

\paragraph{Human Evaluation} While \textsc{Sim} has been shown to correlate with human judgements \citep{mis}, we evaluate whether UAR as an automatic evaluation metric for style transfer accuracy is reflective of human judgements, or if performance is limited to drawing conclusions on the ability to evade automated AID systems only. Human evaluation of this task is challenging as untrained humans are not likely to be able to discriminate nuanced authorship styles easily, where as a neural model like UAR is able to reliably discriminate nuanced authorship styles between thousands of candidate authors simultaneously \citep{luar}. Such a setup of identifying the author from a candidate set of thousands of authors would be too cognitively challenging for untrained human annotators so we alternatively first test their ability on a simpler task. Our task provides example posts from a source author in our dataset as well as example posts from a target author in our dataset and asks annotators to determine which author wrote a post that is randomly selected from one of the two authors. On 675 task instances, with three human annotators per instance, we find humans (77.6\%) underperform UAR (84\%) significantly ($p = 0.05$) at this task. Regardless, we next measure agreement between human and UAR judgements on random style transfer outputs from all of our baselines with the same task setup. On 675 task instances, with three human annotators per instance, we find an agreement coefficient (adjusted for chance) of $0.24$ between human and UAR judgements, indicating ``fair agreement'' \citep{agreement}. We note inter-annotator agreement is only $0.23$ also indicating ``fair agreement'' between the human annotators themselves. On style transfer outputs, we find no significant difference ($p = 0.05$) between humans (63.4\%) and UAR (62.2\%) in discrimination accuracy, making them equally difficult to ``fool" in a binary classification context. Further experimental details about this evaluation can be found in the Appendix. These results indicate some human validation of our evaluation metrics, however, unlike many standard NLP tasks, this task appears to be more difficult for untrained humans than neural models. We believe further validation with expert human annotators, such as forensic linguists, is merited and we leave this as a future direction for researchers with access to such a population to explore.

\subsection{Authorship Obfuscation Experiments}
\label{sec:aid}

To determine \textsc{Styll}'s effectiveness in  providing an adversarial challenge to prior well-established AID models, we perform AID on the style transfer outputs over a large pool of candidate authors ($N$) that includes the source and target authors in the pool amongst many other authors. We recreate the retrieval setup from \citet{luar} with $|N| = 111,396$ authors from the ``test\_target'' split\footnote{UAR Embeddings are trained on the ``train\_*'' splits and our source and target authors are from the ``test\_query'' split.} of the Reddit Million User Dataset (MUD) \citep{accountlinking} using UAR Embeddings to rank users in likelihood of them being the author of a given post with a FAISS index \citep{faiss}. We measure a variety of common retrieval metrics: \textsc{R@8}, \textsc{MRR}, and \textsc{Mean Rank}. We also measure \textsc{Confusion}, a simple percentage of occurrences \textsc{Styll} is able to confuse the AID model into ranking the target author of the style transfer higher than the source author. The results of performing AID can be found in Table \ref{table:uar-retrieval-results}. The \textsc{Confusion} metric demonstrates that \textsc{Styll} is able to confuse the AID model well over 50\% the time on the ``Single'' and ``Diverse'' dataset variants and on the more content-controlled ``Single'' variant, \textsc{Styll} is able to force the AID model to rank the target author within the first 8 results 12-13\% of the time. Importantly, across all variants, we find a substantial drop in the \textsc{R@8} metric to near zero for the source author, indicating difficulty in identifying the source author as a candidate after style transfer. These results demonstrate \textsc{Styll} has promise in preserving the privacy of an author through obfuscation.

\section {Conclusion and Future Directions}
\label{sec:conclusion}

In this work, we study the feasibility of performing and evaluating low-resource authorship style transfer to non-famous authors, a under-studied area of research in text style transfer that has previously largely focused on style transfer to high-resource authors like Shakespeare. We establish datasets, evaluation metrics, and baselines for the task and call for other researchers to further investigate this research direction. Possible future directions include experimenting with larger models and new techniques to reduce hallucination and incoherence in generations and experimenting with this task in a multilingual setting. In Appendix B, we find a major performance benefit as the LLM used with \textsc{Styll} scales ($\text{GPT-2}_{\text{1.5B}}$~$\rightarrow$~$\text{BLOOM}_{\text{7.1B}}$), foreshadowing potential future performance improvement for in-context learning techniques like \textsc{Styll} as larger and more capable LLMs become available.

\paragraph{Data and Resources} We release our datasets, baseline implementations, and code to compute evaluation metrics to encourage further research.

\section*{Ethical Statement}
A broad ethical concern of authorship style transfer research is impersonation. Authorship style transfer, however, can also be used to combat malevolent use of automated authorship identification (AID) systems. For example, in this work, we motivate and experiment with authorship style transfer as a privacy-preserving utility adversarial to automated AID.


\bibliography{aaai24}

\end{document}


\maketitle

\newcommand{\uarembeddingresultstable}{
    \begin{table*}
    \centering
    \setlength{\tabcolsep}{1.8pt}
    \fontsize{9}{11}\selectfont
    \begin{tabular}{lrrrr|rrrr|rrrr}
    \toprule[\heavyrulewidth]
    Method ~~~ & \multicolumn{4}{c|}{Random}& \multicolumn{4}{c|}{Single}& \multicolumn{4}{c}{Diverse}\\
           & \multicolumn{1}{c}{\textsc{Away}} & \multicolumn{1}{c}{\textsc{Towards}} & \multicolumn{1}{c}{\textsc{Sim}} & \multicolumn{1}{c|}{\textsc{Joint}} & \multicolumn{1}{c}{\textsc{Away}} & \multicolumn{1}{c}{\textsc{Towards}} & \multicolumn{1}{c}{\textsc{Sim}} & \multicolumn{1}{c|}{\textsc{Joint}} & \multicolumn{1}{c}{\textsc{Away}} & \multicolumn{1}{c}{\textsc{Towards}} & \multicolumn{1}{c}{\textsc{Sim}} & \multicolumn{1}{c}{\textsc{Joint}} \\ \hline
    $\textsc{Copy}_{\textsc{src}}$   & 0.00 & 0.00 & 1.00 & 0.00 & 0.00 & 0.00 & 1.00 & 0.00 & 0.00 & 0.00 & 1.00 & 0.00 \\
    $\textsc{Copy}_{\textsc{tgt}}$    & 1.00 & 1.00 & 0.00 & 0.00 & 1.00 & 1.00 & 0.00 & 0.00 & 1.00 & 1.00 & 0.00 & 0.00 \\ \hline
    \textsc{Capi}   & 0.42 & 0.02 & 0.89 & 0.17 & 0.56 & 0.01 & 0.93 & 0.07 & 0.41 & 0.01 & 0.87 & 0.08 \\
    \textsc{Cont}   & 0.20 & 0.01 & 0.91 & 0.15 & 0.22 & 0.02 & 0.97 & 0.16 & 0.21 & 0.01 & 0.93 & 0.13 \\
    \textsc{Synm}   & 0.23 & 0.02 & 0.92 & 0.17 & 0.22 & 0.01 & 0.95 & 0.10 & 0.17 & 0.01 & 0.91 & 0.07 \\
    \textsc{Punc}   & 0.23 & 0.02 & 0.93 & 0.24 & 0.25 & 0.02 & 0.97 & 0.19 & 0.26 & 0.02 & 0.90 & 0.18 \\
    \textsc{Emoj}   & 0.27 & 0.04 & 0.93 & 0.25 & 0.29 & 0.06 & 0.95 & 0.27 & 0.27 & 0.02 & 0.93 & 0.17 \\ \hline
    $\textsc{Para}_{\textsc{Neu}}$   & 0.78 & 0.01 & 0.58 & 0.05 & 0.91 & 0.01 & 0.60 & 0.05 & 0.87 & 0.05 & 0.53 & 0.18 \\
    $\textsc{Para}_{\textsc{Div}}$   & 0.75 & 0.01 & 0.69 & 0.10 & 0.91 & 0.02 & 0.71 & 0.11 & 0.83 & 0.04 & 0.70 & 0.18 \\
    \textsc{Ling}   & 0.60 & 0.06 & 0.85 & 0.32 & 0.71 & 0.06 & 0.88 & 0.25 & 0.57 & 0.03 & 0.81 & 0.19 \\
    \textsc{Bert}   & 0.22 & 0.02 & 0.72 & 0.13 & 0.29 & 0.01 & 0.69 & 0.10 & 0.30 & 0.01 & 0.60 & 0.08 \\
    $\textsc{Strap}_{p=0.0}$   & 0.98 & 0.02 & 0.16 & 0.05 & 1.00 & 0.00 & 0.23 & 0.00 & 0.97 & 0.02 & 0.22 & 0.04 \\
    $\textsc{Strap}_{p=0.6}$   & 0.99 & 0.02 & 0.08 & 0.04 & 1.00 & 0.00 & 0.13 & 0.01 & 0.97 & 0.02 & 0.11 & 0.03 \\
    $\textsc{Strap}_{p=0.9}$   & 0.99 & 0.02 & 0.05 & 0.03 & 1.00 & 0.00 & 0.08 & 0.00 & 0.97 & 0.01 & 0.05 & 0.01 \\ \hline
    $\textsc{Styll}_{\text{GPT-3}}$  & 0.78 & 0.07 & 0.45 & 0.23 & 0.91 & 0.11 & 0.48 & 0.29 & 0.87 & 0.12 & 0.44 & 0.30 \\
    $\textsc{Styll}_{\text{BLOOM}}$  & 0.70 & 0.11 & 0.54 & \textbf{0.34} & 0.86 & 0.16 & 0.57 & \textbf{0.40} & 0.76 & 0.12 & 0.58 & \textbf{0.36} \\
    \bottomrule[\heavyrulewidth]
    \end{tabular}
    \caption{Automatic evaluation metrics for our set of baselines over three dataset variants in the UAR Embedding space. Our method \textsc{Styll} outperforms on the proposed \textsc{Joint} metric.}
    \label{table:uar-embedding-results}
    \end{table*}
}

\newcommand{\styleembeddingresultstable}{
    \begin{table*}[h]
    \centering
    \setlength{\tabcolsep}{1.8pt}
    \fontsize{9}{11}\selectfont
    \begin{tabular}{lrrrr|rrrr|rrrr}
    \toprule[\heavyrulewidth]
    Method ~~~ & \multicolumn{4}{c|}{Random}& \multicolumn{4}{c|}{Single}& \multicolumn{4}{c}{Diverse}\\
           & \multicolumn{1}{c}{\textsc{Away}} & \multicolumn{1}{c}{\textsc{Towards}} & \multicolumn{1}{c}{\textsc{Sim}} & \multicolumn{1}{c|}{\textsc{Joint}} & \multicolumn{1}{c}{\textsc{Away}} & \multicolumn{1}{c}{\textsc{Towards}} & \multicolumn{1}{c}{\textsc{Sim}} & \multicolumn{1}{c|}{\textsc{Joint}} & \multicolumn{1}{c}{\textsc{Away}} & \multicolumn{1}{c}{\textsc{Towards}} & \multicolumn{1}{c}{\textsc{Sim}} & \multicolumn{1}{c}{\textsc{Joint}} \\ \hline
    $\textsc{Copy}_{\textsc{src}}$   & 0.00 & 0.00 & 1.00 & 0.00 & 0.00 & 0.00 & 1.00 & 0.00 & 0.00 & 0.00 & 1.00 & 0.00 \\
    $\textsc{Copy}_{\textsc{tgt}}$    & 1.00 & 1.00 & 0.00 & 0.00 & 1.00 & 1.00 & 0.00 & 0.00 & 1.00 & 1.00 & 0.00 & 0.00 \\ \hline
    \textsc{Capi}   & 0.69 & 0.08 & 0.89 & 0.24 & 0.70 & 0.12 & 0.93 & 0.29 & 0.63 & 0.03 & 0.87 & 0.09 \\
    \textsc{Cont}   & 0.23 & 0.03 & 0.91 & 0.15 & 0.22 & 0.02 & 0.97 & 0.12 & 0.26 & 0.01 & 0.93 & 0.06 \\
    \textsc{Synm}   & 0.17 & 0.05 & 0.92 & 0.21 & 0.12 & 0.03 & 0.95 & 0.16 & 0.10 & 0.01 & 0.91 & 0.09 \\
    \textsc{Punc}   & 0.21 & 0.05 & 0.93 & 0.22 & 0.18 & 0.05 & 0.97 & 0.20 & 0.25 & 0.04 & 0.90 & 0.14 \\
    \textsc{Emoj}   & 0.17 & 0.03 & 0.93 & 0.14 & 0.12 & 0.03 & 0.95 & 0.14 & 0.12 & 0.01 & 0.93 & 0.09 \\ \hline
    $\textsc{Para}_{\textsc{Neu}}$   & 0.79 & 0.04 & 0.58 & 0.09 & 0.84 & 0.05 & 0.60 & 0.09 & 0.89 & 0.11 & 0.53 & 0.21 \\
    $\textsc{Para}_{\textsc{Div}}$   & 0.95 & 0.03 & 0.69 & 0.10 & 0.96 & 0.09 & 0.71 & 0.18 & 0.92 & 0.03 & 0.70 & 0.07 \\
    \textsc{Ling}   & 0.74 & 0.13 & 0.85 & \textbf{0.33} & 0.75 & 0.17 & 0.88 & 0.36 & 0.71 & 0.04 & 0.81 & 0.12 \\
    \textsc{Bert}   & 0.15 & 0.05 & 0.72 & 0.14 & 0.18 & 0.04 & 0.69 & 0.13 & 0.20 & 0.02 & 0.60 & 0.08 \\
    $\textsc{Strap}_{p=0.0}$   & 0.96 & 0.04 & 0.16 & 0.06 & 0.96 & 0.15 & 0.23 & 0.16 & 0.95 & 0.04 & 0.22 & 0.04 \\
    $\textsc{Strap}_{p=0.6}$   & 0.97 & 0.04 & 0.08 & 0.04 & 0.96 & 0.16 & 0.13 & 0.12 & 0.95 & 0.03 & 0.11 & 0.02 \\
    $\textsc{Strap}_{p=0.9}$   & 0.96 & 0.04 & 0.05 & 0.03 & 0.96 & 0.17 & 0.08 & 0.09 & 0.96 & 0.03 & 0.05 & 0.01 \\ \hline
    $\textsc{Styll}_{\text{GPT-3}}$  & 0.82 & 0.13 & 0.45 & 0.22 & 0.92 & 0.31 & 0.48 & \textbf{0.39} & 0.93 & 0.23 & 0.44 & \textbf{0.33} \\
    $\textsc{Styll}_{\text{BLOOM}}$  & 0.83 & 0.16 & 0.37 & 0.25 & 0.90 & 0.32 & 0.42 & 0.36 & 0.91 & 0.25 & 0.37 & 0.30 \\
    \bottomrule[\heavyrulewidth]
    \end{tabular}
    \caption{Automatic evaluation metrics for our set of baselines over three dataset variants in the Style Embedding space. Our method \textsc{Styll} outperforms baselines on the ``Single'' and ``Diverse'' variants on the proposed \textsc{Joint} metric. The \textsc{Ling} baseline outperforms on the ``Random'' variant where we expect authorship style transfer to be more undefined or outright impossible as discussed in-depth in Section 3. \textsc{Ling} has downsides as a handcrafted solution we discuss in Section 4, under $\textsc{Ling}$.}
    \end{table*}
}

\newcommand{\uarretrievalresultstable}{
    \begin{table*}
    \centering
    \fontsize{9}{11}\selectfont
    \begin{tabular}{llrrr|rrr|r}
    \toprule[\heavyrulewidth]
    & & \multicolumn{3}{c}{\underline{Source Author}} & \multicolumn{3}{c}{\underline{Target Author}} & \\\
    Dataset~~~ & Method ~~~ & \multicolumn{1}{c}{R@8} & \multicolumn{1}{c}{MRR} & \multicolumn{1}{c}{\textsc{Mean Rank}} & \multicolumn{1}{c}{R@8} & \multicolumn{1}{c}{MRR} & \multicolumn{1}{c}{\textsc{Mean Rank}} & \multicolumn{1}{c}{\textsc{Confusion}} \\ \hline
    \multirow{4}{*}{Random}
    & \textsc{Source}   & 0.73 & 0.55 & 11.67 & 0.00 & 0.00 & 49,553.51 & 0.00 \\
    & \textsc{Target}    & 0.00 & 0.00 & 54,454.28 & 0.60 & 0.43 & 425.93 & 1.00 \\ \cline{2-9} 
    & $\textsc{Styll}_{\text{GPT-3}}$  & 0.00 & 0.00 & 14,000.11 & 0.00 & 0.00 & 27,416.64 & \textbf{0.26} \\ 
    & $\textsc{Styll}_{\text{BLOOM}}$  & 0.07 & 0.03 & 8,913.84 & 0.00 & 0.00 & 25,262.64 & 0.19 \\ \hline
    \multirow{4}{*}{Single}
    & \textsc{Source}   & 0.53 & 0.50 & 46.07 & 0.00 & 0.00 & 10,329.16 & 0.01 \\
    & \textsc{Target}    & 0.00 & 0.00 & 12,211.86 & 0.67 & 0.46 & 518.40 & 0.93 \\ \cline{2-9} 
    & $\textsc{Styll}_{\text{GPT-3}}$  & 0.02 & 0.01 & 9,264.95 & 0.13 & 0.08 & 7,593.85 & \textbf{0.62} \\
    & $\textsc{Styll}_{\text{BLOOM}}$  & 0.00 & 0.00 & 5,816.22 & 0.12 & 0.06 & 4,144.24 & 0.60 \\ \hline
    \multirow{4}{*}{Diverse}
        & \textsc{Source}   & 0.40 & 0.32 & 2,569.87 & 0.00 & 0.00 & 37,122.09 & 0.05 \\
    & \textsc{Target}    & 0.00 & 0.00 & 53,932.73 & 0.67 & 0.49 & 257.93 & 1.00 \\ \cline{2-9} 
    & $\textsc{Styll}_{\text{GPT-3}}$  & 0.04 & 0.01 & 30,989.25 & 0.01 & 0.01 & 24,436.47 & \textbf{0.60} \\
    & $\textsc{Styll}_{\text{BLOOM}}$  & 0.06 & 0.06 & 22,419.18 & 0.00 & 0.00 & 19,515.24 & 0.53 \\
    \bottomrule[\heavyrulewidth]
    \end{tabular}
    \caption{Authorship identification performance with UAR embeddings over $|N| = 111,396$ authors with style transfer outputs from our method. \textsc{Styll} confuses the AID model over 50\% of the time on the ``Single'' and ``Diverse'' variants and forces the target author into the first 8 results 12-13\% of the time on the ``Single'' variant.}
    \label{table:uar-retrieval-results}
    \end{table*}
}

\newcommand{\postiveresultstable}{
    \begin{table*}[ht]
    \centering
    \fontsize{9}{11}\selectfont
    \begin{tabular}{p{5.2cm}p{1.7cm}p{4.2cm}p{4.5cm}}
    \toprule[\heavyrulewidth]
    Target Style Examples & Method & Input $~~\rightarrow~~$ Output & Analysis \\ \hline
        ``CRAP DID I MISS IT" \newline
        ``HES FALLING WITH STYLE!" \newline
        ``LOL DOESNT MATTER THEY SCORED" \newline
        \ldots 13 more examples &
        $\textsc{Styll}_{\text{BLOOM}}$ &
        ``That short throw was nice."$~~\rightarrow$~~\newline
        ``IT WAS A GOOD SHORT THROW" &
        Mimics the capitalization style of the target author and removes end mark punctuation.
    \\ \hline
        ``no way this dude can make this field goal" \newline
        ``lol he sure did" \newline
        ``more like bullet to the head" \newline
        \ldots 13 more examples &
        $\textsc{Styll}_{\text{GPT-3}}$ &
        ``WOOOOOOOOOOO. And the justin fuente era has officially started!"$~~\rightarrow$~~\newline
        ``justin fuente era officially started" &
        Transforms to lower case capitalization style and a terser style.
    \\ \hline
        ``oh crap..." \newline
        ``Sounds about right...  Georiga will likely move up to \#2" \newline
        ``Channeling madden..." \newline
        \ldots 13 more examples &
        $\textsc{Styll}_{\text{BLOOM}}$ &
        ``Lol I'm done already"$~~\rightarrow$~~\newline
        ``I'm done with this... I'm done" &
        Applies the use of ellipses, similar to the target author.
    \\ \hline
        ``Come to Dickson - I'm a man of my word" \newline
        ``You have my love - for the moment" \newline
        \ldots 14 more examples &
        $\textsc{Styll}_{\text{BLOOM}}$ &
        ``Holy shit! There's hope!"$~~\rightarrow$~~\newline
        ``Oh my God - there's hope" &
        Applies the use of an interruptive dash, similar to the target author.
    \\ \hline
        ``The [voyage of Jewish refugees]\newline(https://www.ushmm.org/wlc/en/\newline article.php?ModuleId=10005267) in the [St. Louis](http://www.bbc.com\newline /news/magazine-27373131) exemplifies this fact. \textbackslash n\textbackslash n Great fucking point, u/*******!" \newline
        \ldots 15 more examples &
        $\textsc{Styll}_{\text{GPT-3}}$ &
        ``After going through 4 submissions that broke the peak limiter this was a great one to land on. "$~~\rightarrow$~~\newline
        ``After reviewing [four submissions](https://www.reddit.com/r/\newline gifs/comments/3a0a6j/peak\_ \newline 
        limiter\_review\_4\_submissions\newline 
        \_that/)? this one was a standout" &
        The target author heavily provides links and references in Markdown. Our method follows this stylistic pattern, but the hyperlink it generates is of course a hallucination without the model having real-world knowledge of URLs.
    \\ \hline
        ``\#HEAD ON\#\textbackslash n\textbackslash n\#APPLY DIRECTLY TO THE FOREHEAD\#" \newline
        \ldots 15 more examples &
        $\textsc{Styll}_{\text{GPT-3}}$ &
        ``Happy cake day!\textbackslash n"$~~\rightarrow$~~\newline
        ``\#BIG HAPPY BIRTHDAY\#" &
        Mimics using pairs of "\#" around a capitalized message, a very unique indicator of the target author.
    \\ 
    \bottomrule[\heavyrulewidth]
    \end{tabular}
    \caption{Selected example outputs generated by \textsc{Styll} with analysis. More generations and examples of common failure modes can be found in Appendix H and I.}
    \label{table:positive-results}
    \end{table*}
}

\newcommand{\fullpostiveresultstable}{
    \begin{table*}[h]
    \centering
    \fontsize{9}{11}\selectfont
    \begin{tabular}{p{5.2cm}p{1.7cm}p{4.2cm}p{4.5cm}}
    \toprule[\heavyrulewidth]
    Target Style Examples & Method & Input $~~\rightarrow~~$ Output & Analysis \\ \hline
        ``CRAP DID I MISS IT" \newline
        ``HES FALLING WITH STYLE!" \newline
        ``LOL DOESNT MATTER THEY SCORED" \newline
        \ldots 13 more examples &
        $\textsc{Styll}_{\text{BLOOM}}$ &
        ``That short throw was nice."$~~\rightarrow$~~\newline
        ``IT WAS A GOOD SHORT THROW" &
        Mimics the capitalization style of the target author and removes end mark punctuation.
    \\ \hline
        ``no way this dude can make this field goal" \newline
        ``lol he sure did" \newline
        ``more like bullet to the head" \newline
        \ldots 13 more examples &
        $\textsc{Styll}_{\text{GPT-3}}$ &
        ``WOOOOOOOOOOO. And the justin fuente era has officially started!"$~~\rightarrow$~~\newline
        ``justin fuente era officially started" &
        Transforms to lower case capitalization style and a terser style.
    \\ \hline
        ``they keep hillbillies from getting dented? *sweet*" \newline
        ``of course, this is a *thing*" \newline
        ``*Never* forget your towel!" \newline
        \ldots 13 more examples &
        $\textsc{Styll}_{\text{GPT-3}}$ &
        ``SHUT UP PORTUGAL"$~~\rightarrow$~~\newline
        ``Please be *quiet*, Portugal" &
        Applies the use of asterisks (Markdown formatting) for emphasis instead of capitalization, similar to the target author.
    \\  \hline
        ``oh crap..." \newline
        ``Sounds about right...  Georiga will likely move up to \#2" \newline
        ``Channeling madden..." \newline
        \ldots 13 more examples &
        $\textsc{Styll}_{\text{BLOOM}}$ &
        ``Lol I'm done already"$~~\rightarrow$~~\newline
        ``I'm done with this... I'm done" &
        Applies the use of ellipses, similar to the target author.
    \\ \hline
        ``Come to Dickson - I'm a man of my word" \newline
        ``You have my love - for the moment" \newline
        \ldots 14 more examples &
        $\textsc{Styll}_{\text{BLOOM}}$ &
        ``Holy shit! There's hope!"$~~\rightarrow$~~\newline
        ``Oh my God - there's hope" &
        Applies the use of an interruptive dash, similar to the target author.
    \\ \hline
        ``The [voyage of Jewish refugees]\newline(https://www.ushmm.org/wlc/en/\newline article.php?ModuleId=10005267) in the [St. Louis](http://www.bbc.com\newline /news/magazine-27373131) exemplifies this fact. \textbackslash n\textbackslash n Great fucking point, u/*******!" \newline
        \ldots 15 more examples &
        $\textsc{Styll}_{\text{GPT-3}}$ &
        ``After going through 4 submissions that broke the peak limiter this was a great one to land on. "$~~\rightarrow$~~\newline
        ``After reviewing [four submissions](https://www.reddit.com/r/\newline gifs/comments/3a0a6j/peak\_ \newline 
        limiter\_review\_4\_submissions\newline 
        \_that/)? this one was a standout" &
        The target author heavily provides links and references in Markdown. Our method follows this stylistic pattern, but the hyperlink it generates is of course a hallucination without the model having real-world knowledge of URLs.
    \\ \hline
        ``\#HEAD ON\#\textbackslash n\textbackslash n\#APPLY DIRECTLY TO THE FOREHEAD\#" \newline
        \ldots 15 more examples &
        $\textsc{Styll}_{\text{GPT-3}}$ &
        ``Happy cake day!\textbackslash n"$~~\rightarrow$~~\newline
        ``\#BIG HAPPY BIRTHDAY\#" &
        Mimics using pairs of "\#" around a capitalized message, a very unique indicator of the target author.
    \\
    \bottomrule[\heavyrulewidth]
    \end{tabular}
    \caption{Selected example outputs generated by \textsc{Styll} with analysis.}
    \label{table:full-positive-results}
    \end{table*}
}

\newcommand{\negativeresultstable}{
    \begin{table*}[h]
    \centering
    \fontsize{9}{11}\selectfont
    \begin{tabular}{p{5.2cm}p{1.7cm}p{4.2cm}p{4.5cm}}
    \toprule[\heavyrulewidth]
    Target Style Examples & Method & Input $~~\rightarrow~~$ Output & Analysis \\ \hline
        ``CRAP DID I MISS IT" \newline
        ``HES FALLING WITH STYLE!" \newline
        ``LOL DOESNT MATTER THEY SCORED" \newline
        $\ldots$ 13 more examples &
        $\textsc{Styll}_{\text{BLOOM}}$ &
        ``McMillan sighting!"$~~\rightarrow$~~\newline
        ``McMillan got a look." &
        A low-quality paraphrase results in ``sighting'' being paraphrased to ``got a look'' which is a misinterpretation.
    \\ \hline
        ``Are they still talking about Daredevil over there?" \newline
        ``So that's what having X-ray vision kick in must be like." \newline
        ``That's more the sign on the bathroom door the next morning." \newline
        $\ldots$ 13 more examples &
        $\textsc{Styll}_{\text{GPT-3}}$ &
        ``Hey man, my wife and I lived with my mom for a bit until we got back on track. Nothing wrong with coming home for a while."$~~\rightarrow$~~\newline
        ``Hey, it's okay to come home for a little while." &
        Style transfer results in some information from the original source text being lost.
    \\ \hline
        ``Yes, yes it does." \newline
        ``Tennessee looks like trash, tbh." \newline
        ``These are two of the worst offenses in America." \newline
        $\ldots$ 13 more examples &
        $\textsc{Styll}_{\text{GPT-3}}$ &
        ``That UConn game though"$~~\rightarrow$~~\newline
        ``That UConn game was terrible." &
        Hallucination during generation results partial new information not present in the original source text. In this particular example, we suspect the hallucination may be influenced by the overall negative tone the target author uses.
    \\ \hline
        ``What is your take on gun control?" \newline
        ``I like how the maggot part is what's wrong with that statement." \newline
        ``I was never interested in Warhammer until this comment." \newline
        $\ldots$ 13 more examples &
        $\textsc{Styll}_{\text{BLOOM}}$ &
        ``Eh? Definitely bot copying Crysis."$~~\rightarrow$~~\newline
        ``What is that? What? What? I don't know what I'm talking about." &
        Hallucination during generation results in a generation that is unrecognizably different from the source text.
    \\
    \bottomrule[\heavyrulewidth]
    \end{tabular}
    \caption{Example outputs representative of the common failure modes in the generation produced by \textsc{Styll}.}
    \end{table*}
}

\newcommand{\ablationuarresultstable}{
    \begin{table*}[h]
    \centering
    \setlength{\tabcolsep}{1.8pt}
    \fontsize{9}{11}\selectfont
    \begin{tabular}{lrr|rr|rr}
    \toprule[\heavyrulewidth]
    Method ~~~ & \multicolumn{2}{c|}{Random}& \multicolumn{2}{c|}{Single}& \multicolumn{2}{c}{Diverse}\\
           & \multicolumn{1}{c}{\textsc{Confusion}} & \multicolumn{1}{c|}{\textsc{Joint}} & \multicolumn{1}{c}{\textsc{Confusion}} & \multicolumn{1}{c|}{\textsc{Joint}} & \multicolumn{1}{c}{\textsc{Confusion}} & \multicolumn{1}{c}{\textsc{Joint}} \\ \hline
    - \textsc{Des}  & 0.00 & \textbf{0.18} & 0.00 & \textbf{0.33} & 0.22 & \textbf{0.44} \\
    + \textsc{Des}  & \underline{0.11} & 0.10 & \underline{0.33} & 0.29 & \underline{0.33} & 0.38 \\
    \bottomrule[\heavyrulewidth]
    \end{tabular}
    \caption{Automatic evaluation metrics over smaller dataset variants of 3 source authors and 3 target authors for our method \textsc{Styll} ablating the use of style descriptors in the UAR Embedding space. \textsc{Des} indicates usage of style descriptors in the few-shot style transfer prompt described in step 3 of Section 4.}
    \end{table*}
}

\newcommand{\strapvsstylltable}{
    \begin{table*}[h]
    \centering
    \fontsize{9}{11}\selectfont
    \begin{tabular}{p{3.5cm}p{5cm}p{4.5cm}}
    \toprule[\heavyrulewidth]
    Target Style Examples & Input $~~\rightarrow~~$ \textsc{Strap} Output & Input $~~\rightarrow~~$ \textsc{Styll} Output \\ \hline
        ``Who wants salad for lunch?" \newline
        ``Dirt road?  Extra tire squeal for you!" \newline
        ``Divorce can ruin a person financially despite their steady income." \newline
        \ldots 13 more examples &
        ``Who are pretty much ``buying" the development from Aalto."$~~\rightarrow$~~\newline
        ``who's buying A who's buying Aal who's buying Aalto's development pretty much" &
        ``Who are pretty much ``buying" the development from Aalto."$~~\rightarrow$~~\newline
        ``Aalto is getting financial support from a variety of sources in order to develop their project."

    \\ \hline
        ``If you take my hat, I'm taking some of your teeth" \newline
        ``If you would die tomorrow, what do you want to eat?"\newline
        ``Be. The ball." \newline
        \ldots 13 more examples &
        ``Lots of misconceptions about airbag deployment. It's not necessarily the point of impact that determines whether the bags deploy or not, it is the deceleration of the occupants as determined by the airbag system. Caravans are relatively soft so the impact was likely not severe enough."$~~\rightarrow$~~\newline
        ``The airbag is notthe airbag .The airbag is not the airbag is not the airbag is not the airbag is not the airbag is not the airbag" &
        ``Lots of misconceptions about airbag deployment. It's not necessarily the point of impact that determines whether the bags deploy or not, it is the deceleration of the occupants as determined by the airbag system. Caravans are relatively soft so the impact was likely not severe enough."$~~\rightarrow$~~\newline
        ``There are a lot of myths about airbag deployment. One of them is that it is always based on the impact. The truth is that it is based on the deceleration of the occupants. Caravans are not as stiff as other types of vehicles, so the impact may have been less severe." \newline
    \\ \hline
        ``anyone have the full source?" \newline
        ``You're a idiot."\newline
        ``No beta? No way." \newline
        \ldots 13 more examples &
        ``Nice to see Madoka there enjoying it with her."$~~\rightarrow$~~\newline
        ``. . The first thing is the first thing is the first thing is the first"
        &
        ``Nice to see Madoka there enjoying it with her."$~~\rightarrow$~~\newline
        ``It's refreshing to see Madoka having fun."
    \\
    \bottomrule[\heavyrulewidth]
    \end{tabular}
    \caption{Example outputs generated by $\textsc{Strap}_{p=0.0}$  and $\textsc{Styll}_{\text{GPT-3}}$ on the ``Diverse'' dataset. Examples are chosen at random as to be a representative sample of general output quality. Not only does \textsc{Styll} outperform \textsc{Strap} on the \textsc{Joint} automatic evaluation metric as discussed in Section 6, in our low-resource setting we qualitatively observe that its generated outputs are of far higher quality and fluency. \textsc{Strap} greatly suffers with issues of generation repetition and coherence when trained with such little training data.}
    \end{table*}
}

\newcommand{\llmuarresultstable}{
    \begin{table*}[h]
    \centering
    \setlength{\tabcolsep}{1.8pt}
    \fontsize{9}{11}\selectfont
    \begin{tabular}{lrrrr|rrrr|rrrr}
    \toprule[\heavyrulewidth]
    Method ~~~ & \multicolumn{4}{c|}{Random}& \multicolumn{4}{c|}{Single}& \multicolumn{4}{c}{Diverse}\\
           & \multicolumn{1}{c}{\textsc{Away}} & \multicolumn{1}{c}{\textsc{Towards}}& \multicolumn{1}{c}{\textsc{Sim}} & \multicolumn{1}{c|}{\textsc{Joint}} & \multicolumn{1}{c}{\textsc{Away}} & \multicolumn{1}{c}{\textsc{Towards}}& \multicolumn{1}{c}{\textsc{Sim}} & \multicolumn{1}{c|}{\textsc{Joint}} & \multicolumn{1}{c}{\textsc{Away}} & \multicolumn{1}{c}{\textsc{Towards}}& \multicolumn{1}{c}{\textsc{Sim}} & \multicolumn{1}{c}{\textsc{Joint}} \\ \hline
    $\text{GPT-2}_{\text{1.5B}}$  & 0.92 & 0.17 & 0.10 & 0.17 & 0.97 & 0.13 & 0.13 & 0.16 & 0.93 & 0.14 & 0.14 & 0.17 \\
    $\text{GPT-3}_{\text{6.7B}}$  & 0.78 & 0.07 & 0.45 & 0.23 & 0.91 & 0.11 & 0.48 & 0.29 & 0.87 & 0.12 & 0.44 & 0.30 \\
    $\text{GPT-J}_{\text{6B}}$  & 0.71 & 0.11 & 0.56 & \textbf{0.35} & 0.87 & 0.15 & 0.60 & 0.39 & 0.77 & 0.12 & 0.60 & \textbf{0.36} \\
    $\text{OPT}_{\text{6.7B}}$  & 0.74 & 0.13 & 0.43 & 0.31 & 0.89 & 0.14 & 0.44 & 0.32 & 0.82 & 0.13 & 0.50 & 0.35 \\
    $\text{BLOOM}_{\text{7.1B}}$  & 0.70 & 0.11 & 0.54 & 0.34 & 0.86 & 0.16 & 0.57 & \textbf{0.40} & 0.76 & 0.12 & 0.58 & \textbf{0.36} \\
    $\text{FLAN-T5}_{\text{3B}}$  & 0.69 & 0.05 & 0.61 & 0.21 & 0.88 & 0.06 & 0.62 & 0.22 & 0.77 & 0.06 & 0.63 & 0.26 \\
    \bottomrule[\heavyrulewidth]
    \end{tabular}
    \caption{Automatic evaluation metrics for our method \textsc{Styll} applied with different large language models for step 3 of Section 4 over three dataset variants in the UAR Embedding space.}
    \end{table*}
}

\newcommand{\llmstyleresultstable}{
    \begin{table*}[h]
    \centering
    \setlength{\tabcolsep}{1.8pt}
    \fontsize{9}{11}\selectfont
    \begin{tabular}{lrrrr|rrrr|rrrr}
    \toprule[\heavyrulewidth]
    Method ~~~ & \multicolumn{4}{c|}{Random}& \multicolumn{4}{c|}{Single}& \multicolumn{4}{c}{Diverse}\\
           & \multicolumn{1}{c}{\textsc{Away}} & \multicolumn{1}{c}{\textsc{Towards}}& \multicolumn{1}{c}{\textsc{Sim}} & \multicolumn{1}{c|}{\textsc{Joint}} & \multicolumn{1}{c}{\textsc{Away}} & \multicolumn{1}{c}{\textsc{Towards}}& \multicolumn{1}{c}{\textsc{Sim}} & \multicolumn{1}{c|}{\textsc{Joint}} & \multicolumn{1}{c}{\textsc{Away}} & \multicolumn{1}{c}{\textsc{Towards}}& \multicolumn{1}{c}{\textsc{Sim}} & \multicolumn{1}{c}{\textsc{Joint}} \\ \hline
    $\text{GPT-2}_{\text{1.5B}}$  & 0.91 & 0.20 & 0.10 & 0.13 & 0.92 & 0.25 & 0.13 & 0.15 & 0.93 & 0.18 & 0.14 & 0.14 \\
    $\text{GPT-3}_{\text{6.7B}}$  & 0.82 & 0.13 & 0.45 & 0.22 & 0.92 & 0.31 & 0.48 & 0.39 & 0.93 & 0.23 & 0.44 & 0.33 \\
    $\text{GPT-J}_{\text{6B}}$  & 0.77 & 0.17 & 0.56 & 0.31 & 0.87 & 0.29 & 0.60 & \textbf{0.43} & 0.86 & 0.24 & 0.60 & \textbf{0.41} \\
    $\text{OPT}_{\text{6.7B}}$  & 0.79 & 0.19 & 0.43 & 0.29 & 0.90 & 0.28 & 0.44 & 0.34 & 0.88 & 0.25 & 0.50 & 0.38 \\
    $\text{BLOOM}_{\text{7.1B}}$  & 0.74 & 0.19 & 0.54 & \textbf{0.32} & 0.84 & 0.30 & 0.57 & 0.40 & 0.83 & 0.26 & 0.58 & \textbf{0.41} \\
    $\text{FLAN-T5}_{\text{3B}}$  & 0.81 & 0.08 & 0.61 & 0.19 & 0.91 & 0.14 & 0.62 & 0.25 & 0.87 & 0.10 & 0.63 & 0.22 \\
    \bottomrule[\heavyrulewidth]
    \end{tabular}
    \caption{Automatic evaluation metrics for our method \textsc{Styll} applied with different large language models for step 3 of Section 4 over three dataset variants in the Style Embedding space.}
    \end{table*}
}

\newcommand{\decodinguarresultstable}{
    \begin{table*}[h]
    \centering
    \setlength{\tabcolsep}{1.8pt}
    \fontsize{9}{11}\selectfont
    \begin{tabular}{lrrrr|rrrr|rrrr}
    \toprule[\heavyrulewidth]
    Method ~~~ & \multicolumn{4}{c|}{Random}& \multicolumn{4}{c|}{Single}& \multicolumn{4}{c}{Diverse}\\
           & \multicolumn{1}{c}{\textsc{Away}} & \multicolumn{1}{c}{\textsc{Towards}} & \multicolumn{1}{c}{\textsc{Sim}} & \multicolumn{1}{c|}{\textsc{Joint}} & \multicolumn{1}{c}{\textsc{Away}} & \multicolumn{1}{c}{\textsc{Towards}} & \multicolumn{1}{c}{\textsc{Sim}} & \multicolumn{1}{c|}{\textsc{Joint}} & \multicolumn{1}{c}{\textsc{Away}} & \multicolumn{1}{c}{\textsc{Towards}} & \multicolumn{1}{c}{\textsc{Sim}} & \multicolumn{1}{c}{\textsc{Joint}} \\ \hline
    $t = 0.1$  & 0.67 & 0.06 & 0.65 & 0.24 & 0.85 & 0.11 & 0.66 & 0.32 & 0.75 & 0.08 & 0.68 & 0.31 \\
    $t = 0.7$  & 0.70 & 0.11 & 0.54 & 0.34 & 0.86 & 0.16 & 0.57 & \textbf{0.40} & 0.76 & 0.12 & 0.58 & \textbf{0.36} \\
    $t = 0.9$  & 0.73 & 0.15 & 0.42 & \textbf{0.35} & 0.87 & 0.18 & 0.47 & 0.38 & 0.79 & 0.15 & 0.46 & 0.35 \\
    \bottomrule[\heavyrulewidth]
    \end{tabular}
    \caption{Automatic evaluation metrics for our method \textsc{Styll} applied with different settings of \texttt{temperature} for step 3 of Section 4 over three dataset variants in the UAR Embedding space.}
    \end{table*}
}


\appendix
\onecolumn

\section{Implementation and Experiment Details}
\label{sec:implementation-details}
\begin{itemize}
    \item For our ``Single" dataset variant: ``r/AskReddit'' is the most common subreddit in our Reddit corpus. However, since the posts in AskReddit can be about a wide-range of topics such as science, politics, video games, etc., we chose to draw users who exclusively posted in the second most common subreddit in the dataset, which happened to be ``r/CFB''.
    \item For our manual handcrafted baselines: Tokenization and part-of-speech tagging was done using spaCy (\texttt{en\_core\_web\_trf-3.3.0}) \citep{spacy}, synonyms were found using WordNet \citep{wordnet} and NLTK \citep{nltk}, and Python package \texttt{pycontractions} is used to manipulate contractions. Swapped words are transformed to match the inflection and case of the original word with the package \texttt{lemminflect}.
    \item For our \textsc{Bert} baseline: We use the \texttt{roberta-base} \citep{roberta} model. Tokens with the part-of-speech tags ``AUX'', ``ADP'', and ``PART'' are never swapped. 
    \item For our LLM baselines: Prompts for step 1 and 2 were manually tested against Reddit posts from ``r/ELI5'' and ``r/AskReddit'' from 2022 not included in our Reddit corpus, we find them to be robust across many LLMs (see Appendix \ref{sec:llm}). We use these models with a \texttt{temperature} setting of 0.7 and a \texttt{top\_p} setting of 1.0, the default setting for GPT-3.
    \item For our human evaluation: Our population is drawn from student volunteers from a natural language processing course at a university, a population slightly more trained than the average crowd worker. We utilize the Amazon MTurk platform for annotation and have each task instance annotated by three separate annotators. For our first and second human evaluations, we find an inter-annotator agreement  of $0.23$ and $0.23$ respectively \citep{krippendorff}.
\end{itemize}

\section{{\textsc{Styll} Performance with Large Language Models}}
\label{sec:llm}

Depending on the specific dataset variant, various large language models perform the best or are competitive. In this paper, we choose to select $\text{BLOOM}_{\text{7.1B}}$ out of these models as the open source and reproducible model to demonstrate \textsc{Styll}, however, other models of similar scale perform competitively. 

~\\
\llmuarresultstable
\llmstyleresultstable

OPT was explicitly trained on Reddit data \citep{opt}, possibly including some of our source author and target author posts. Other models were trained on variants of Common Crawl data. To the best of our knowledge, using various search tools 
available\footnote{\url{https://c4-search.apps.allenai.org/}}\footnote{\url{https://huggingface.co/spaces/bigscience-data/roots-search}}, we were unable to find instances of our Reddit data in the pre-training corpora of other models. Common Crawl, as a breadth-first crawler, only contains a small amount of Reddit data, often a few thousand pages per month according to the Common Crawl Index Server\footnote{\url{https://index.commoncrawl.org/}}, a miniscule fraction of the monthly submission volume. Any Reddit posts that may exist in pre-training corpora would not be explicitly given as examples of the source author's style and would be only sparsely represented in the pre-training copora compared to the style of high-resource authors like Shakespeare that appear in pre-training corpora frequently.

\section{Style Transfer with ChatGPT}
\label{sec:chatgpt}

Anecdotally, we have seen ChatGPT \citep{chatgpt}  be able to mimic the style of various online authors within in its training set (such as commenters on HackerNews\footnote{\url{https://news.ycombinator.com/item?id=33860562}}). While ChatGPT was not available at the time of these experiments, ChatGPT is a fine-tuned version of the GPT-3 model we use in this paper using Reinforcement Learning from Human Feedback to optimize for conversation. As future work, we believe it may be useful to investigate whether this process allows ChatGPT to perform style transfer any better than the standard GPT-3 model.

\section{\textsc{Styll} Style Descriptors Ablation}
\label{sec:ablations}

While not using style descriptors (\textsc{Des}) with $\textsc{Styll}_{\text{GPT-3}}$ performs best on the \textsc{Joint} metric (bolded), using style descriptors helps boost \textsc{Confusion} (underlined), which is a percentage measurement of how often we are closer to the target author than the source author in the embedding space. The \textsc{Joint} metric equally weighs style transfer accuracy and meaning preservation, however, we may choose to manually select for higher style transfer accuracy in our approach, sacrificing some meaning preservation. In the authorship style transfer task, this can be desirable as we discuss in Section 5, under $\textsc{Joint}(s, t)$. In our work, \textsc{Styll} refers to performing style transfer with the procedure described in Section 4 with the use of style descriptors, closely following the prompt format in \citet{styletransferrecipe}.

~\\
\ablationuarresultstable

\section{Effect of Decoding Parameters on \textsc{Styll}}
\label{sec:decoding}
In this paper, we use the default decoding parameters (\texttt{temperature} setting of 0.7 and a \texttt{top\_p} setting of 1.0)  as our decoding parameters for all experiments. Curious readers may ask what effect decoding parameters have on \textsc{Styll}. We provide the results for $\textsc{Styll}_{\text{BLOOM}}$ with various \texttt{temperature} settings.

\decodinguarresultstable
\raggedbottom
\pagebreak

\section{\textsc{Styll} Results in Style Embedding Space}
\label{sec:strap-style-emb-results}
\styleembeddingresultstable
\raggedbottom
\pagebreak

\section{\textsc{Strap} vs. \textsc{Styll} Outputs}
\label{sec:strap-vs-styll}
\strapvsstylltable
\raggedbottom
\pagebreak

\section{\textsc{Styll} Selected Generations}
\label{sec:styll-full-positive-results}
\fullpostiveresultstable
\raggedbottom
\pagebreak

\section{\textsc{Styll} Common Failure Modes}
\label{sec:styll-common-failure-modes}
\negativeresultstable
\raggedbottom
\pagebreak

\raggedbottom
\pagebreak


\section{Resources}
\label{sec:resources}

\noindent We provide links and citations to resources used in this paper which provide license information, documentation, and their intended use. Our usage follows the intended usage of all resources.

~\\ ~\\ \noindent We utilize the following models:
\begin{itemize}
    \item GPT-2 \citep{gpt2}
    \item $\text{GPT-3}_{\text{6.7B}}$ (\texttt{text-curie-001}) \citep{gpt3}
    \item $\text{GPT-J}_{\text{6B}}$ \citep{gptj}
    \item $\text{OPT}_{\text{6.7B}}$ \citep{opt}
    \item $\text{BLOOM}_{\text{7.1B}}$ \citep{bloom}
    \item $\text{FLAN-T5}_{\text{3B}}$ \citep{flant5}
    \item \textsc{Strap} \citep{strap}
    \item Mutual Implication Score \citep{mis}
    \item Learning Universal Authorship Representations (UAR) Embedding model \citep{luar}
    \item Style Embedding model \citep{styleemb}
    \item RoBERTa (\texttt{roberta-base}) \citep{bert,roberta}
\end{itemize}

~\\ \noindent We utilize the following datasets:
\begin{itemize}
    \item Shakespeare author imitation dataset \citep{shakespeare}
    \item Reddit Million User Dataset \citep{lir}
    \item CoLA dataset \citep{cola}
    \item \textsc{paranmt-50m} dataset \citep{paranmt,strap}
    \item WordNet \citep{wordnet}
\end{itemize}

~\\ \noindent We utilize the following software:
\begin{itemize}
    \item Transformers \citep{transformers}
    \item Sentence-Transformers \citep{sentencetransformers}
    \item FAISS \citep{faiss}
    \item spaCy (\texttt{en\_core\_web\_trf-3.3.0}) \citep{spacy}
    \item NLTK \citep{nltk}
    \item \texttt{pycontractions} \textendash~ \url{https://pypi.org/project/pycontractions/}
    \item \texttt{lemminflect} \textendash ~ \url{https://pypi.org/project/lemminflect/}
\end{itemize}

~\\ \noindent We estimate the total compute budget and detail computing infrastructure used to run the computational experiments found in this paper below:
\begin{itemize}
    \item 1x NVIDIA A100 Tensor Core GPU / 100GB RAM / 4x CPU -- 305 hours
\end{itemize}
\raggedbottom
\pagebreak

\bibliography{aaai24}